\documentclass{article}

\usepackage{arxiv}

\usepackage[utf8]{inputenc} 
\usepackage[T1]{fontenc}    
\usepackage{hyperref}       
\usepackage{url}            
\usepackage{booktabs}       
\usepackage{amsfonts}       
\usepackage{nicefrac}       
\usepackage{microtype}      
\usepackage{graphicx}
\usepackage{natbib}
\usepackage{doi}
\usepackage{float}
\usepackage{amsmath}
\usepackage{multirow}
\usepackage{algorithm}
\usepackage{algpseudocode}

\newcommand{\beginsupplement}{%
        \setcounter{table}{0}
        \renewcommand{\thetable}{S\arabic{table}}%
        \setcounter{figure}{0}
        \renewcommand{\thefigure}{S\arabic{figure}}%
     }

\setcitestyle{numbers,square}

\title{Masked Mixers for Language Generation and Retrieval}

\author{ \href{https://orcid.org/0000-0003-1661-4579}{\includegraphics[scale=0.06]{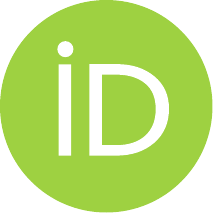}\hspace{1mm}Benjamin L. Badger}\thanks{The author would like to thank Seek AI for support during the research and writing of this manuscript. Much of this work was performed while at Guidehouse. Code may be found on \url{https://github.com/blbadger/maskedmixers}.} \\
	Seek AI \\
	\texttt{badger@seek.ai} \\
}

\date{}

\renewcommand{\headeright}{Technical Report}
\renewcommand{\undertitle}{Technical Report}

\hypersetup{
pdftitle={Masked Mixers for Language Generation and Retrieval},
pdfsubject={Deep Learning},
pdfauthor={Benjamin L. Badger},
pdfkeywords={Deep Learning, Representation Theory},
}

\begin{document}
\maketitle

\begin{abstract}
    Attention mechanisms that confer selective focus on a strict subset of input elements are nearly ubiquitous in language models today. We posit there to be downside to the use of attention: most input information is lost. In support of this idea we observe poor input representation accuracy in transformers and more accurate representation in what we term masked mixers, which replace self-attention with masked convolutions.  The masked mixer learns causal language modeling more efficiently than early transformer implementations and even outperforms optimized, current transformers when training on small ($n_{ctx}<512$) but not larger context windows. Evidence is presented for the hypothesis that differences in transformer and masked mixer training efficiencies for various tasks are best predicted by input representation accuracy, or equivalently global invertibility.  We hypothesize that the information loss exhibited by transformers would be more detrimental to retrieval than generation, as the former is more closely approximated by a bijective and thus invertible function. We find that masked mixers are more effective retrieval models both when the pretrained embedding model is unchanged as well as when the embedding model is modified via cosine similarity-based InfoNCE loss minimization.  A small masked mixer is shown to outperform a large and near state-of-the-art transformer-based retrieval model, despite the latter being trained with many orders of magnitude more data and compute.
\end{abstract}

    \begin{center}
    \begin{table}[H]
    \begin{center}
    \renewcommand{\arraystretch}{1.2}
    \begin{tabular}{||l c c c c c c c c c||} 
     \hline
      Model & $n=32$ & 64 & 128 & 256 & 512 & 1024 & 2048 & 4096 & 8192 \\ [0.5ex] 
     \hline\hline
      Transformer & 95.0 & 92.3 & 88.9 & 84.5 & 79.1 & 72.6 & 65.7 & 57.6 & 50.4 \\
     \hline
      e5 Mistral Instruct & 95.1 & 93.3 & 90.9 & 88.5 & 85.6 & 82.2 & 78.4 & 73.5 & 68.8  \\ 
     \hline
      Masked Mixer & \textbf{98.2} & \textbf{97.2} & \textbf{95.8} & \textbf{93.4} & \textbf{90.2} & \textbf{86.5} & \textbf{81.7} & \textbf{76.6} & \textbf{70.6} \\ 
     \hline
    \end{tabular}
    \end{center}
    \vspace{0.1cm}
    \caption{FineMath retrieval top-1 accuracy (\%) for $n$ sample sizes, each denoting retrieval with one query, one matching target and $n-2$ nonmatching targets.}
    \label{table10}
    \end{table}
    \end{center}


\section{Introduction}

    Since the introduction of the transformer \citep{vaswani2017attention}, many deep learning application domains have witnessed a rapid increase in the state-of-the-art as a result of incorporation of this architecture. Originally designed for language processing tasks such as machine translation, the transformer has been found to be well suited to increases in model parameters such that ever-larger models may be trained on ever-larger datasets with continued increases in various metrics of goodness \citep{radford2018improving, le2023bloom, touvron2023llama, dubey2024llama}. The use of attention without recurrent mechanisms proved to be extremely influential in the sequence modeling field and beyond and has led to the introduction of other modules that attempt to address constraints inherent in dot-product attention \citep{gu2024mambalineartimesequencemodeling}.

    In spite of this impressive success, it is less clear how efficiently transformers or other models using attention may be trained for various language tasks: many types of architectures experience increases in goodness with an increased parameter numbers and dataset sizes, making early claims that the transformer scaled in superior ways to other models somewhat dubious.  One way to investigate this question indirectly is to observe how information passes from the input to various hidden layers of the model, which can be considered to be closely related to the question of how these hidden layers represent the model's input.  Current investigations on this topic often focus on the method by which transformers transfer information to and from different tokens (cf. \citep{olsson2022context}), but we focus on a simpler question: \textit{how much} rather than \textit{how} information is passed from the input to various parts of the model. In particular, we focus on the information present in the model's last hidden layer, as this is most commonly used for tasks such as causal modeling (with a language modeling head applied) and retrieval.

    The goal of any attention operation is to focus on some input elements at the expense of others. Here we use the term `focus' to attempt to avoid overloading `attention' with too many meanings, and define the term as analogous to `weight' or the amount a change in an input element affects a change in the output. Equivalently, any effective attention transformation is by definition not globally invertible because out-of-focus input elements may be modified without a change in the output. 
    
    Transformers used for language generation consist of many sequential self-attention modules (separated by MLPs) each with multiple attention heads per module, however, such that the restricted focus inherent in each attention transformation can no longer necessarily be ascribed to the model as a whole. In light of observations that trained language models tend to require high numerical precision from a tiny subset of all weights and activations which are often large in absolute value \citep{dettmers2022gpt3, dettmers20228bit} we hypothesize that the inherent information-limiting characteristic of attention does translate to the transformer model as a whole. Conversely, we hypothesize that a model substituting linear transformations for attention would not exhibit these informational limitations.

    In this work we observe that transformer model hidden layers do not contain sufficient uniquely identifying information to allow for accurate input reconstruction via gradient-based input representation algorithms, whereas the introduced `masked mixer' models with self-attention substituted for masked convolutions do and are in that sense globally invertible. We then compare causal language training efficiencies for a number of modeling tasks and find that the more a task requires something similar to what would be termed model invertibility, the more the masked mixer tends to outperform the transformer for both small and larger datasets.  In general we corroborate the observations of original vision-based MLP mixer work in finding that attention is superfluous to much of the transformer's performance \citep{tolstikhin2021mlp, melas2021you}.

    We investigate the performance of masked mixers and transformers for language retrieval task where a query to text matching task. We introduce a novel approach to training an accessory retrieval model using the embeddings of a CLM-pretrained model, and find that the masked mixer's embeddings result in far more accurate retrieval. We then investigate the ability of masked mixers and transformers to perform retrieval when the embedding model is trained directly via InfoNCE-based cosine similarity loss, and find that once again masked mixers are substantially more accurate than transformers in this paradigm. We conclude with the observations that a masked mixer exhibits more accurate retrieval than a large and near state-of-the-art transformer-based retrieval model pretrained with approximately 10,000x the compute, as well as scaling data to show that transformers require many orders of magnitude more retrieval data to match the masked mixer's accuracy.

\subsection{Related Work}

    The body of work on deep learning representation is large. Much work towards understanding transformers has revolved around explanations for how attention transformations are capable of adding, removing, and transferring information from one token or patch to another (see for example \citep{olsson2022context} in the context of one-headed attention). For vision models, methods for applying gradient descent in order to observe representations in a model's hidden layers were pioneered by such studies as \citep{mahendran2015understanding}. Notably the method of representation employed in this work is in some ways much more straightforward than there: instead of performing smoothing and gradient orientation procedures to gain a comprehensible representation, we perform vanilla gradient descent followed by pseudoinversion to recover tokens.
    
    The substitution of self-attention for convolutions in the transformer was introduced independently by \citep{tolstikhin2021mlp} and \citep{melas2021you} as the MLP-Mixer, a vision model designed primarily for image classification tasks. The authors of the former work noted that for a given fixed-sized image dataset the MLP-Mixer performed slightly worse than vision transformers, but do not directly compare their efficiencies for fixed compute training runs \citep{tolstikhin2021mlp}. We modify the mixer architecture introduced in these works in a few notable ways for causal language modeling: we apply triangular masking directly to the convolutional (ie the sequence MLP) weight parameter matrices both during training and inference, replace the mixer layer's two sequential linear transformations separated by an activation layer on the sequence dimension with a single masked convolution with no activation, and optionally use non-unitary convolutional kernels or multi-headed convolutions.
    
    A language application of the MLP mixer based on Bloom-filtered inputs was introduced in \citep{fusco2022pnlp} but was not designed for causal language modeling, and presents a number of architectural decisions that make this model unsuitable for autoregressive language modeling tasks that transformers are commonly trained for. 

    Elaborations on the use of convolutions in place of attention have met with success in attempts to model language with sub-quadratic complexity in the sequence dimension. In \citep{fu2023monarchmixersimplesubquadratic}, Fu and colleagues replace attention with Monarch matrix-based sequence transformations for efficient sub-quadratic language modeling, but motivated by that goal the authors use a Monarch matrix-specific causal parametrization rather than parameter masking for causal language modeling. Poli and colleagues \citep{poli2023hyenahierarchylargerconvolutional} introduced the Hyena architecture consisting of a recurrent structure with long convolutions combined with a gating operation with the same goal, and also do not perform direct masking. We comment that these works provide valuable insight into achieving near-linear complexity language modeling via efficient convolutional operations, but our approach to efficient language modeling focuses on global invertibility predicting training efficiency rather than sequence complexity and thus we proceed with an entirely different theoretical framework and do not investigate sub-quadratic complexity convolutional operations (Monarch or Toeplitz weight matrices etc.) which trade inference efficiency for training inefficiencies in terms of loss achieved per compute applied.

    The finding that information often passes linearly between tokens in simplified transformers \citep{olsson2022context} and the observation that purely linear deep models (ie with no nonlinearities between matrix multiplications) exhibit non-linear learning characteristics even if they lack the representational power of nonlinear models \citep{Saxe_2019} suggest that simple inter-token linear transformations (for example, 1-dimensional convolutions) may capture much of the dynamics of more complex inter-token transformations, such as those in self-attention modules.

    To appreciate the motivation behind this work it is vital to understand that accurate input representation is equivalent to global invertibility \textit{only within the feasible input domain}, not any arbitrary input. This is important to appreciate because it explains how models that are composed of non-invertible transformations (such as every linear layer in this work) can exhibit global invertibility, and is significant departure from work focused on converting model transformations to invertible versions of those transformations \citep{zha2021invertibleattention}.

\section{Our Contribution}

    We introduce the following:
    \begin{enumerate}
    \item An input representation method for language models to test for global invertibility
    \item The masked mixer, a modification of the MLP mixer for language modeling
    \item A bidirectional modeling implementation with higher throughput than BERT-style masked language modeling
    \item An non-trivial autoencoding method for sequence models with promising retrieval characteristics
    \item A high-throughput retrieval sampling method and architecture in which a model is trained to match embeddings
    \item Masked mixer adaptations for cosine similarity-based InfoNCE retrieval training
    \end{enumerate}

    The primary insight behind this work is that differences in a model's input representation accuracy, or equivalently its global invertibility, are capable of predicting a wide range of differences in training efficiency and performance between transformers and models of identical architecture except for the substitution of attention for linear transformations.  Using this principle as a guide, we introduce a new foundational model architecture that far surpasses the performance of the transformer for retrieval and autoencoding.

\section{Accurate Self and Non-Self Token Representation in Masked Mixers}

    How much input information does a model contain in its hidden layers? More specifically, given some hidden layer values $O_l(a, \theta)$ can we regenerate the model's input $a$ or in other words can we globally invert the model? We use a gradient descent-based approach to address this question for language models. With respect to information that can be used to uniquely identify a language input, the answer for transformers is not very much. We introduce the masked mixer and show that this model accurately represents its inputs before and after training, and even accurately represent a limited number of non-self tokens.

    Properly speaking, autoregressive language models such as transformers exist as multifunctions during training as there are many input elements and many output elements, but are used as (many-to-one) functions during inference as there is one next token output generated per forward pass. We investigate invertibility in both settings.

    \subsection{Input representation background}
    
    In this work we measure the information present in a model's hidden layer representation of the input by attempting to recover the input using gradient descent on an initially random input, minimizing a chosen metric on the activations of that hidden layer as described in \citep{badger2023depth}. The goal is to invert a model such that information in the hidden layer's activations (which may be thought of as that layer's representation) is used to identify the model's input. 
    
    Briefly, for some input $a$ and some chosen layer $l$ of model $\theta$ such that $O_l(a, \theta)$ is a vector space, gradient descent is performed on the norm of the difference between the hidden layer activations given some initially random input $a_0 = \mathcal{N}(a, \mu=1/2, \sigma=1/20)$ and its activations given $a$, where the only values that that can be modified are the elements of the initially random input $a_n$. For example, this procedure using the $L^1$ metric is used in Equation (\ref{eq1}).
    
    \begin{equation}
    a_{n+1} = a_n - \eta * \nabla_{a_n} ||O_l(a_n, \theta) - O_l(a, \theta)||_1 \\
    \label{eq1}
    \end{equation}

    We use a scheduled learning rate $\eta$ that decreases linearly from $\eta$ to $\eta / 10$ as $n \to N$ (where the iteration number $n$ is increased until it reaches the final number of iterations $N$, which we set somewhat arbitrarily to $N=500$) which empirically results in the fast optimization with a reasonable $N$. 
    
    Language models typically operate using discrete inputs, usually integer tokens. This means that one cannot immediately apply (\ref{eq1}) but must instead either convert the input tokens to a differentiable vector space or else optimize some other value and convert that value to the discrete input $a_N$. Elsewhere it was found experimentally that the latter process tends to be more stable across a wide range of inputs \citep{badgersentence}, and that is the approach we use for this work.

    To be specific, we optimize the embeddings of the input tokens rather than the input itself. The embeddings may be computed by matrix-vector multiplication using the embedding weight matrix $W$ such that $e = Wa$. With an initially random embedding $e_0 = \mathcal{N}(e, \mu=1/2, \sigma=1/20)$ we perform gradient descent on the embedding using an $L^1$ metric on the layer in question as denoted in (\ref{eq2}).

    \begin{equation}
    e_{n+1} = e_n - \eta * \nabla_{e_n} ||O_l(e_n, \theta) - O_l(e, \theta)||_1 \\
    \label{eq2}
    \end{equation}

\begin{figure}[h]
        \centering
        \includegraphics[width=0.9\textwidth]{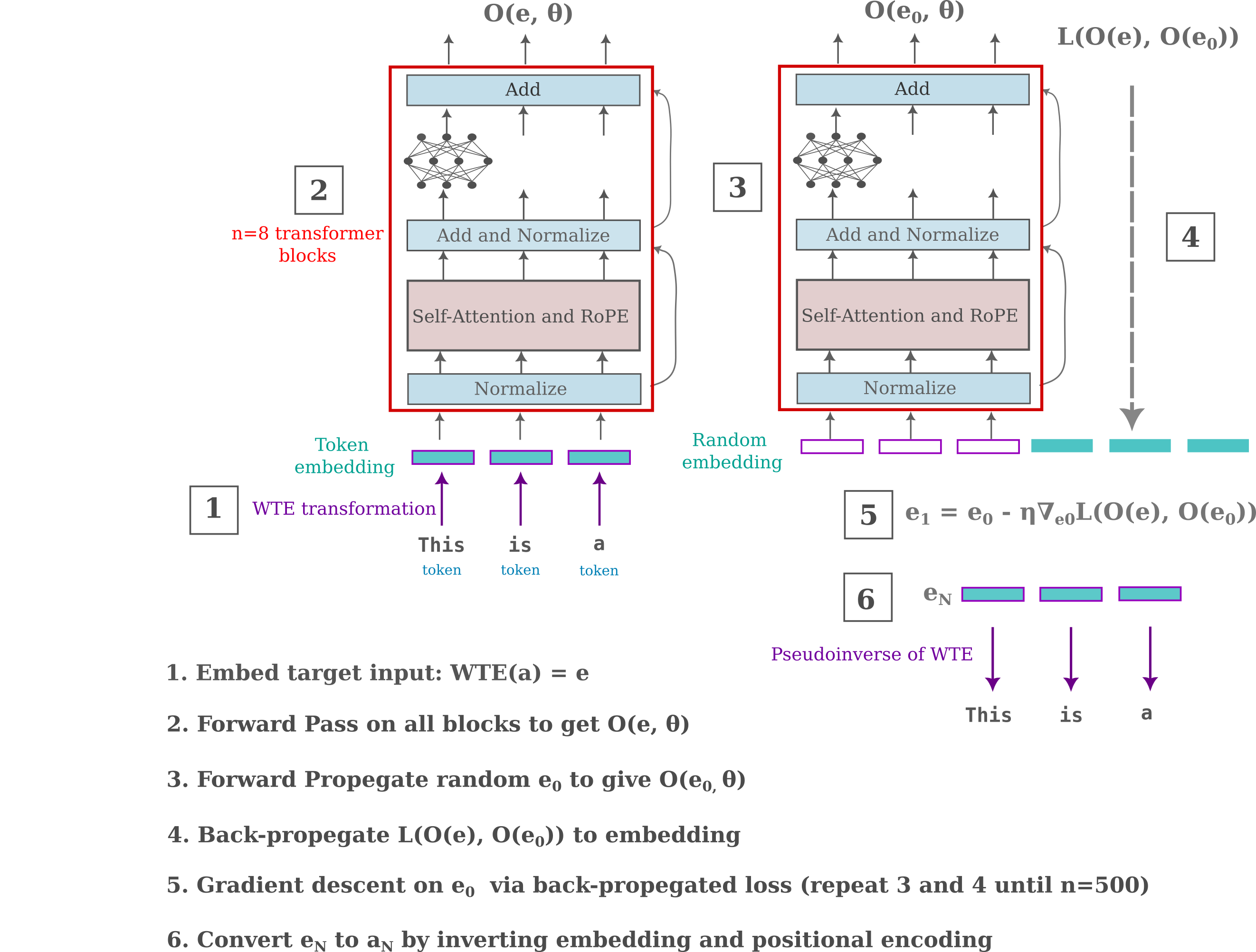}
        \caption{Indirect input representation method applied to a Llama-style transformer.}
        \label{fig1}
    \end{figure}

    In general the embedding weight matrix $W$ is non-square and non-invertible. $e_N$ is therefore converted back to the generated input $a_N$ via left multiplication by the Moore-Penrose pseudo-inverse $W^+$ via (\ref{eq3}), which is a form of generalized matrix inverse defined by Equation (\ref{eq4}). If positional encoding is applied to the embedding, it is subtracted before (\ref{eq3}) is applied. This process is illustrated in Figure \ref{fig1} for convenience.

    \begin{equation}
    a_N = W^+e_N
    \label{eq3}
    \end{equation}
    
    \begin{equation}
    W^+ = \lim_{\alpha \to 0^+} (W^T W + \alpha I)^{-1} W^T
    \label{eq4}
    \end{equation}
    
    If the gradient is effectively backpropegated to the input $e_n$ in Equation (\ref{eq2}) then $\vert \vert O_l(e_N, \theta) - O_l(e, \theta) \vert \vert_1 < \epsilon$ for some small $\epsilon$. We establish the value of $\epsilon$ in order to check this condition by adding a very small amount of Gaussian noise to the initial input embedding before observing the metric distance of the output of this noised input to $a$ such that $\epsilon = \vert \vert O_l(e + {N}(e, \mu=0, \sigma=1/20), \theta)) - O_l(e, \theta) \vert \vert_1$. This check allows us to confirm that iterations of (\ref{eq2}) results in sufficient minimization of the metric on the output (in this case $L^1$) such that the corresponding $a$ and $a_N$ are nearly equivalent for the model. The particular amount of Gaussian noise to add in order to specify the output shift $\epsilon$ was determined in part by observing that the chosen value leads to no token changes upon pseudoinversion, meaning that typically $W^+e = W^+(e + {N}(e, \mu=0, \sigma=1/20))$ although this depends somewhat on the transformation via the word-token embedding weights $W$.

    \subsection{Masked mixer architecture}

    Elsewhere it was observed that large and otherwise capable language models (Llama-2 7b and 70b etc.) exhibit relatively inaccurate input representation for all but the shallowest few hidden layers \citep{badgersentence}. It was also observed that vision MLP mixer models (\citep{melas2021you, tolstikhin2021mlp}) have superior input representation accuracy for non-self tokens compared to vision transformers \citep{badgervits}, leading us to wonder if a mixer adapted to causal language modeling could exhibit this accurate input representation. We hypothesized that adapting an MLP-mixer to the process of language generation would give us a model suitable for autoregressive language generation and other tasks, but with much more accurate input representation abilities.

    We adapt the MLP-Mixer architecture for causal language modeling as follows: first all 1D convolutions (ie `MLPs' on the sequence dimension) are reshaped and lower-triangular masked such that only inputs from tokens $t_0, t_1, ..., t_{n-1}$ have non-zero weights for the token $t_n$. As for causal language modeling (CLM) -style transformer models, output values are also shifted such that the loss compares $O(t_n)$ and $t_{n+1}$ for all $n \in N$ such that the model may be trained on all tokens of an input with one forward and backward pass.

   The details of the triangular masking process are as follows: first the 1D convolution weights are reshaped to place the model dimension and number of tokens in the last two tensor dimensions, a triangular mask is then applied to those weights, and finally the convolutional weight data is re-written with the masked weight data, reverted to its original shape. A succinct PyTorch \citep{NEURIPS2019_bdbca288} implementation of this convolutional weight masking (for any convolutional kernel size) using einops \citep{rogozhnikov2022einops} and torch.tril lower triangular masking is as follows:
    
\begin{verbatim}
masked_conv = tril(rearrange(conv.weight, 'f d k -> k f d'))
conv.weight.data = rearrange(masked_conv, 'k f d -> f d k').contiguous()
\end{verbatim}

    where `f' and `d' are both equal to the number of tokens in the context window and `k' is the 1-dimensional kernel size.  A 1D convolution with a kernel size of 1 can also be implemented using `torch.linear' on the sequence dimension in which the triangular weight masking is a more pithy `conv.weight.data = torch.tril(conv.weight)' after reshaping, but experimentally this does not exhibit as much throughput as the 1D convolutional kernel method above nor is it as flexible for non-unitary kernels.
    
\begin{figure}[h]
        \centering
        \includegraphics[width=0.75\textwidth]{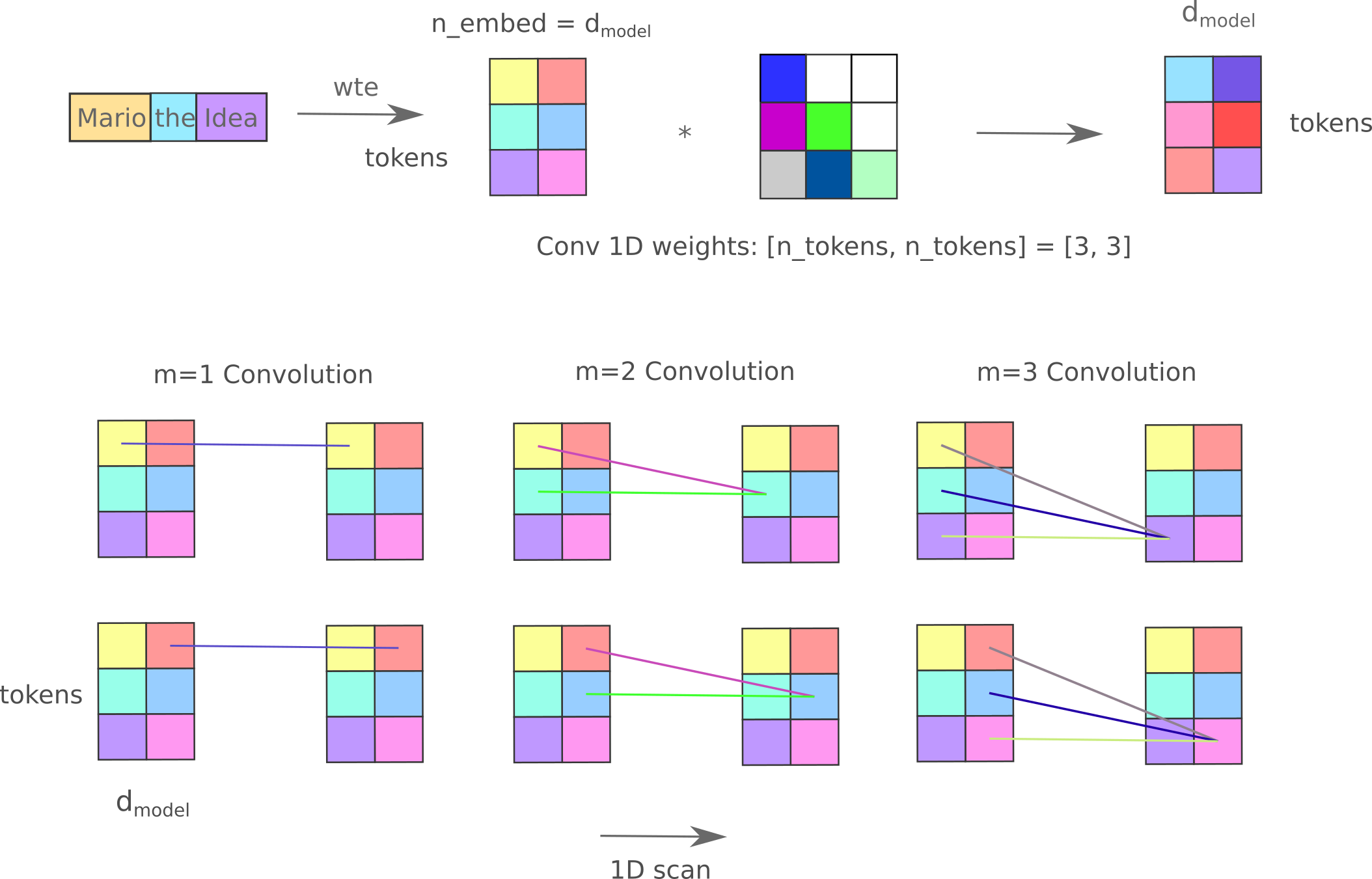}
        \caption{Causal language modeling via masking convolutional weights.}
        \label{fig2}
    \end{figure}

    A diagrammatic interpretation of how these operations act to enforce causal language modeling is shown in Figure \ref{fig2}. The primary difference between these operations and the method used to CLM masking during during transformer training is that in that case a mask is usually applied to a self-attention module's activations, whereas here the mask is applied to the weights directly.

    The forward pass is nearly identically to all-next-token training for transformers with some slight modification for reshaping. The use of the same loss function, tokenizer, and context window allows us to compare directly between transformer and masked mixer losses in order to assess training efficiency for these two models.
    
    We call these models `Masked Mixers' to emphasize 1) that the mask is an intrinsic part of the model for both training and inference and 2) these mixers in some cases no longer resemble MLPs, as they contain convolutions with non-unitary kernels or projected multi-headed kernels.

    \subsection{Masked mixers but not transformers exhibit accurate input representations}

    We now turn to the task of measuring input representation accuracy, and we introduce a method for achieving this goal. Notably, we observe input representation before and after causal language model training, because the training procedure can drastically change a model's representation accuracy. 

    In the spirit of an information theoretic approach, we modify the normalized Hamming metric for comparing input representations to their corresponding inputs. Normalization and further modification is required because language inputs are not of fixed size, but masked mixers require fixed context windows as currently implemented.
    
    We define this metric as follows: given input $x$ and generated representation of that input $y$ such that each element of $x$ is an integer corresponding to an element in the tokenizer set $\{t_m\}$, which may be stated precisely as shown in Equation \ref{eq51}, then the normalized Hamming metric is given in Equation (\ref{eq5}). In words, this metric is the fraction of indices of the input where the generated representation's token does not match the input's token while ignoring the input's pad token positions. The smaller the normalized Hamming distance, the more similar the input is to the model's representation and the more information that representation carries. We measure before training as well as during and after 12 hours of RTX 3060 training on the first 2M examples of the TinyStories \citep{eldan2023tiny} dataset.
    
    \begin{equation}
        x = (x_1, x_2, ..., x_n), \; y = (y_1, y_2, ..., y_n) \in \{0, 1, ..., t_m\}^n
        \label{eq51}
    \end{equation}

    \begin{equation}
        h(x, y) = \frac{1}{n} \mathrm{Card}(\{x_i \neq y_i\}) : x_i \neq t_{pad}
    \label{eq5}
    \end{equation}
    
\begin{figure}[h]
        \centering
        \includegraphics[width=0.8\textwidth]{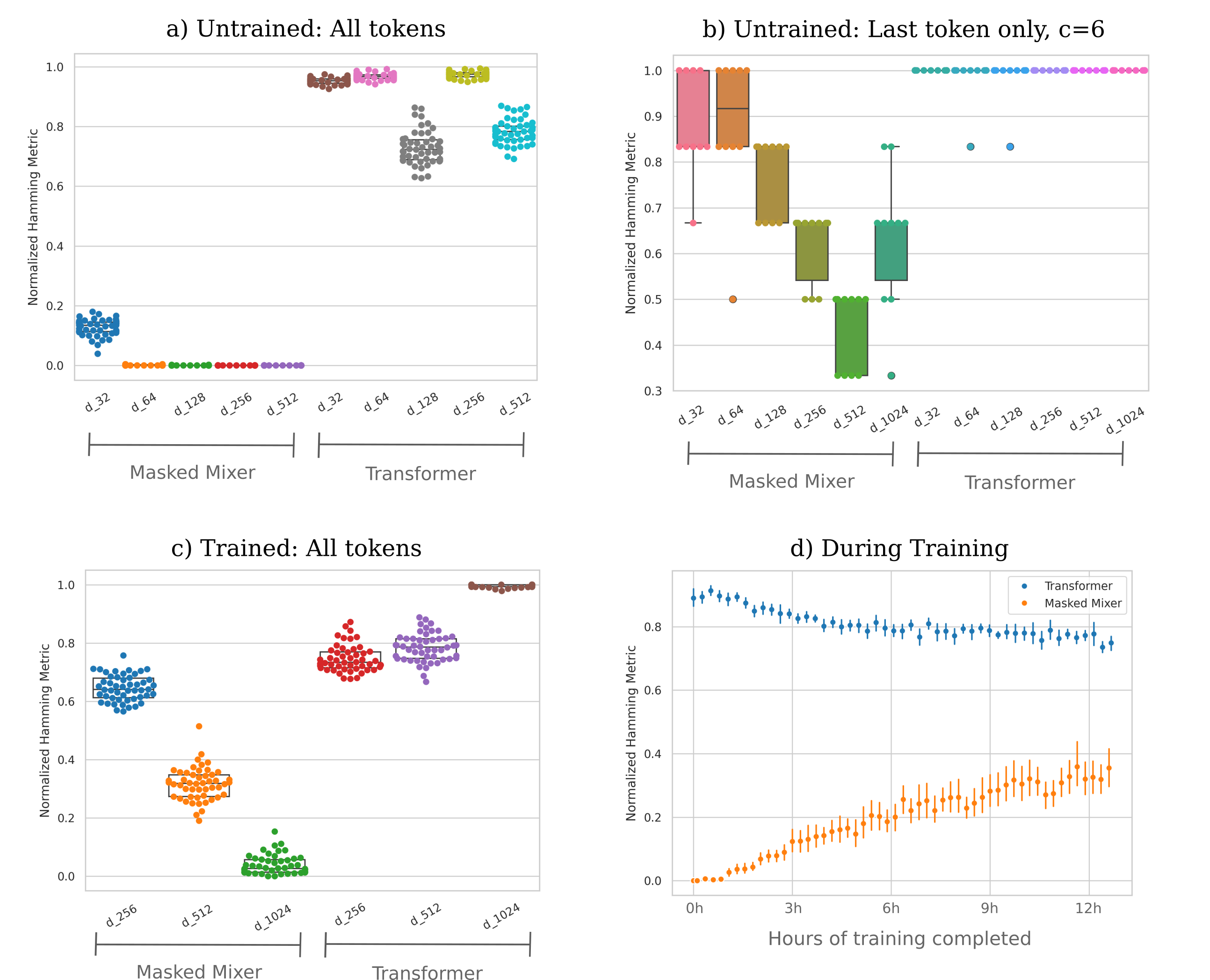}
        \caption{Masked mixers exhibit more accurate input representation than transformers. All models are $n_l=8$ and all transformers are llama architectures with $n_h=32$. In d) the transformer is $d_m=256$ and mixer $d_m=512$, trained on TinyStories. Inputs are random samples from TinyStories.}
        \label{fig3}
    \end{figure}

    As observed elsewhere for larger language models, small transformers designed for TinyStories modeling exhibit very poor input representation such that the information present in the last hidden layer is capable of recovering little if any of an input when (\ref{eq2}) is applied (Figure \ref{fig3}), although this improves somewhat with training. Masked mixers exhibit near-perfect representation before training, and larger models retain this characteristic even after training on Tinystories (Figure \ref{fig3}). 

    Masked mixers also exhibit much more accurate input representation than transformers when gradient descent is performed using only the last hidden layer of the last token (Figure \ref{fig3}b), indicating that they pass more information between tokens than transformers. Neither transformers nor masked mixers exhibit accurate non-self token representation for larger context windows ($c=512$ etc.) if only the last token's last hidden layer is used for backpropagation, indicating that there are limits to the inter-token information bandwidth in mixers as well as transformers before training. 

    The original MLP-mixer architecture employs two sequential convolutional operations between each input token, with an expansion factor (usually set to two) for the layer between convolutions. We find that models with only one convolution operation between tokens (a `flat' masked mixer) have superior inter-token information transfer (Figure \ref{figs3}), but that their self-token representation power is similar.

    Masked mixers of a sufficient $d_{model}$ are in some sense biased towards accurate input representation as untrained models nearly always exhibit perfect input representation regardless of their actual randomly initialized values. In this vein, transformers of a sufficient $d_{model}$ are biased against accurate input representation. It is interesting therefore to observe that masked mixer and transformer input representation abilities converge somewhat during training (Figure \ref{fig3}d), which suggests that for the TinyStories dataset there is some range of optimal input representation accuracies that is associated with minimization of all-next-token prediction during causal language modeling. This convergence is very slow, however, especially for larger models such that the model type's initial bias determines most behavior after training as well (Figure \ref{fig3}c). 
    
    If we scale up the compute to increase the number of TinyStories samples by a factor of 10x we find that indeed the gap in the Hamming metric between masked mixer and transformer is substantially reduced (Figure \ref{figs4}), as is true when training is performed on a much larger and more diverse dataset with >10x the compute (Figure \ref{figs5}). Notably, masked mixers trained on those large datasets experience a general decrease in representational accuracy as the context window size of the training inputs increases (Figure \ref{figs5}).

\section{Masked Mixer and Transformer Causal Language Modeling Efficiency}

    The ability of large transformer models to provide useful causal language modeling without accurate input representation suggests that accurate input representation is not necessary for that task. We reasoned that although input representational accuracy does not necessarily provide greater language modeling capabilities, it could perhaps lead to more efficient training in certain cases.

    To begin to test the training efficiency of the Masked Mixer compared to the Transformer, we employ the same two datasets as for representation studies: the first train using 2M training examples of TinyStories with all evaluation examples with a Llama-2 tokenizer of size 4096 trained on that dataset, where models are trained using a correspondingly small fixed compute budget of 12 hours on an Nvidia RTX 3060, or 2.25 hours on a 4x Nvidia V100 server. We then train on a much larger and more varied dataset, specifically a 10 billion token subset of a extensively refined of recent Common Crawl snapshots called the FineWeb-edu \citep{penedo2024finewebdatasetsdecantingweb} (hereafter termed `FineWeb' for simplicity) with ten times the compute amount, or approximately 20 hours on a 4x V100 server, with 200k steps taken per training run. 

    As the Llama -style transformer uses the same Cross-Entropy Loss function (with loss masking on pad token outputs) on the language modeling head as the Masked Mixer with the same hyperparameters (a context window of 512 unless otherwise noted, AdamW optimizer \citep{loshchilov2019}, etc.), we can directly compare the loss achieved after the fixed compute quantity has been applied, around $4 \times 10^19$ Tensor FLOPs for a 20 hour training run on the 4x V100 server.

    \subsection{Architecture optimizations}

    We perform extensive experiments to optimize the masked mixer's architecture using TinyStories, and find that the use of input representation is an effective guide to training efficiencies for this model. Most notably, we modify the original MLP Mixer architecture to use only one convolution and no nonlinear activation between sequence elements as this was found to exhibit superior input representation accuracy and more efficient training (cf. Section \ref{representation_optim}). The number of layers, hidden layer width, batch sizes, and hyperparameters like learning rates were then optimized on the same dataset via independent line searches. We found that multi-headed masked mixers and masked mixers with non-unitary convolutional kernel sizes were no more efficient, and Softmax-transformed convolutional weight mixers were substantially less efficient, than the standard one-masked-convolution architecture for TinyStories CLM training (Section \ref{multiheaded_mixers}).

    We found that the optimized masked mixer exhibited more efficient TinyStories training characteristics than non-optimized Llama-style transformers (Section \ref{unoptimized_transformers}) but considered this to be an unfair comparison and proceeded to optimize the transformer model's architecture. We found that the default self-attention head number was much too large for optimal training on this dataset and reduced the default (32 heads) to four (Section \ref{optimized_transformers}). As for the masked mixer, we also optimized the number of hidden layers, layer widths, batch sizes, and learning rates using the same dataset. Once this was performed, it was found that the optimized Llama-style transformer was somewhat more efficient for CLM training on TinyStories than the optimized masked mixers.

    As a general note, most comparisons between masked mixers and transformers in this work are performed between models with the same number of layers (usually $n_l=8$ or $n_l=16$) but with $d_m=512$ layer width for transformers and $d_m=1024$ for masked mixers. This difference in $d_m$ is the result of the optimizations detailed above: masked mixers require much less device memory as they contain fewer inter-token parameters (Tables \ref{tables10} and \ref{tables11}), and as such these models are approximately equivalent with respect to on-device memory during training. During the aforementioned optimizations, we find empirically that increasing the transformer's width to $d_m=1024$ with a decrease in batch size results in worse training efficiency and likewise decreasing the masked mixer's width to $d_m=512$ while increasing the batch size leads to worse efficiency. We consider this model-dependent optimization as vital to valid CLM efficiency comparisons, as otherwise one can simply choose architectural or hyperparameter constraints that favor one or the other model. We open our optimization results open to the public in this work and urge other researchers to do the same, as the failure to perform at least some reasonable amount of architectural or hyperparameter optimization leads to relatively meaningless efficiency comparisons between architectures.

    \subsection{Autoregressive inference for masked mixers}

    Masked mixers cannot be inferenced in a manner analogous to transformers because these models have a fixed feedforward context window size, at least without modification (convolutional weight restrictions for example) to the present implementation. This means that a full forward pass is required for each token generation such that the masked mixer inferences with time complexity $O(n^3 * d)$ where $n$ is the number of tokens and $d=d_{model}$. In practice masked mixers inference only slightly slower than transformers that use key-value caching and are thus $O(n^2 * d)$ because of the much lower constant factors for the mixers.

    This architecture requires a new inference method: rather than applying a trained model to a padded input to infer the last character, we instead apply a causal language mask the input just as is done during training and simply infer each next token at the start of the mask. Note that the causal language model mask must be retained during inference to prevent information from right-positioned tokens from influencing the next token prediction at the current token.

    We introduce a simple inference method that makes use of all the information that the masked mixers learns per token position during training. Recall that the mixer has effectively a fixed positional encoding intrinsic in the model's convolutional weights, and that these weights are responsible for all of the inter-token information flow. One can make use of this positional encoding by iterating through the context indices while adding the token generated from the mixer's output at each index. A Python implementation of this approach is as follows:

\begin{verbatim}
for i in range(n_tokens_to_generate, 1, -1):
    loss, output = model(tokens)
    out_token = torch.topk(output, dim=1, k=1).indices.flatten()[-i]
    tokens[..., -i+1] = out_token
\end{verbatim}

    Using this inferencing method we find that masked mixers generate similarly readable outputs to TinyStories prompts as transformers at the same test cross-entropy loss.  For example, given the prompt

    \noindent\fbox{%
    \parbox{\textwidth}{%
    One day, a little boy named Tim went to play with his friend, Sam. They wanted to play a game with a ball. The game was to see who could get the best score. Tim threw the ball and Sam tried to catch it. Sam missed the ball and it went far away. Sam was angry. He said, "I don't like this game. I want to lie down and rest. "Tim said, "Okay, let's lie down and rest. Then we can try again to get a good score." They lay down under a big tree and looked at the sky. They saw birds flying and clouds moving. After a while, they got up and
    }%
}
    a masked mixer yields a fairly coherent story completion:
    
    \noindent\fbox{%
    \parbox{\textwidth}{%
        played a game of catch. Tim threw the ball to Sam, and Sam caught it. They laughed and played until the sun went down.
        At the end of the day, Tim and Sam were tired but happy. They went home and took a nap. They dreamed of playing catch again tomorrow. And they did.
    }%
}

    whereas a transformer with a slightly higher loss gives the following:
    
    \noindent\fbox{%
    \parbox{\textwidth}{%
        started to play again. They played for a long time and had lots of fun. They laughed and played until the sun went down. Then, they went home and told their mom and dad about their game.
        Tim and Sam learned that it is important to be patient and wait for good things to happen. They had a fun day playing together.
    }%
}
    \subsection{Masked mixers are more efficient causal language modelers for small (<512 tokens) but not large-context training}

    To test the ability of masked mixers to be trained efficiently for causal language modeling on datasets that better represent those used for large frontier models while maintaining a reasonable training time on our limited compute, we turned to a 10 billion token subset of the FineWeb-edu \citep{penedo2024finewebdatasetsdecantingweb}. We train a Llama-3 \citep{dubey2024llama} tokenizer of size 8000 on this dataset before using this tokenizer to train randomly-initialized transformers and mixers.

\begin{figure}[h]
        \centering
        \includegraphics[width=0.99\textwidth]{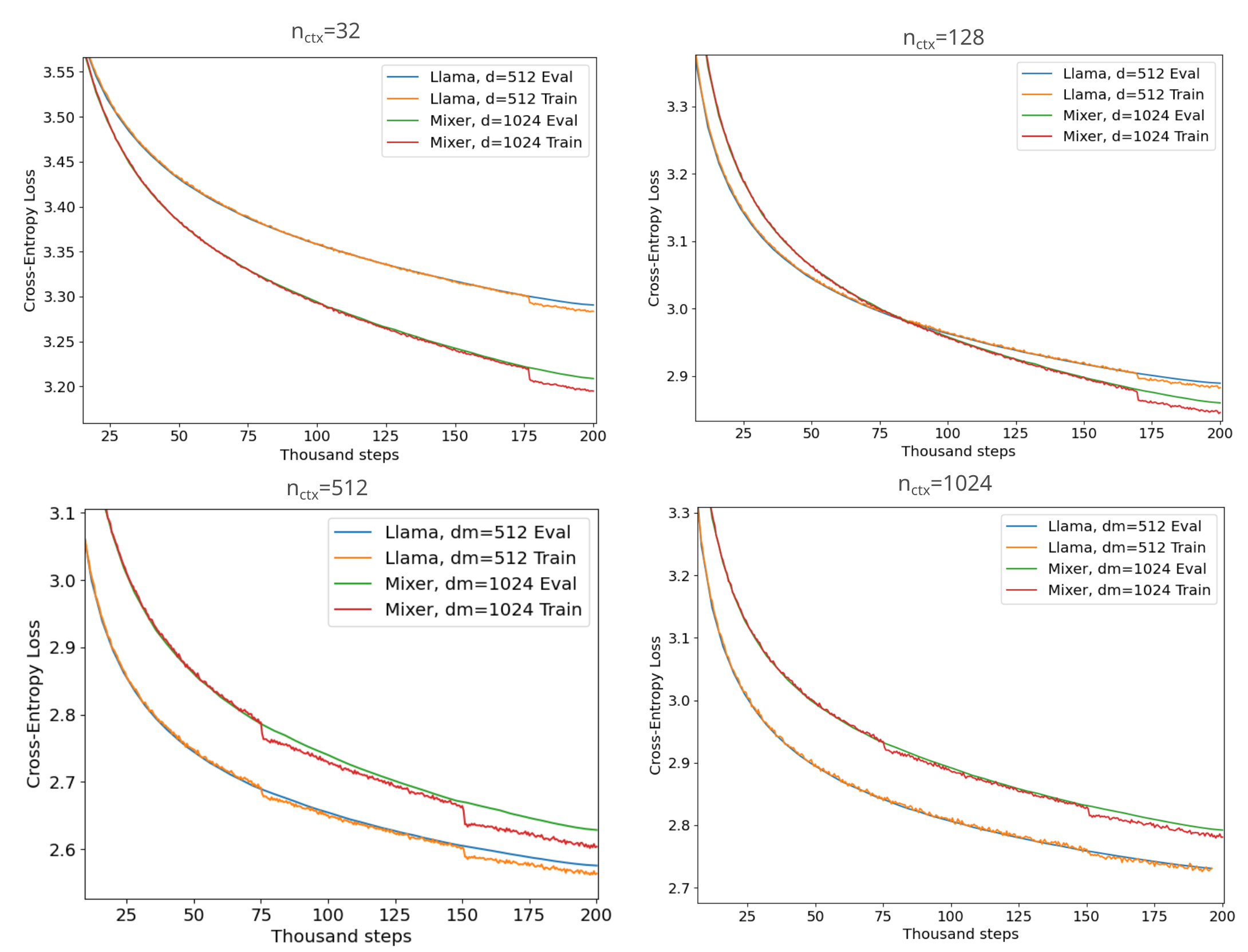}
        \caption{FineWeb Causal Language Model Training.}
        \label{fig13}
    \end{figure}

    We first profiled the Llama transformer model to determine the optimal number of attention heads for this larger dataset, and found that, as for TinyStories, a 4-headed model achieved the lowest per-step loss and required the least time per step (Figure \ref{figs7}). We next trained transformers and masked mixers on the FineWeb dataset with varying context windows and batch sizes according to the rule $n_{ctx} * n_b = 65536$ to prevent memory overflow, and the results of these experiments are shown in Figure \ref{fig13}. We observe that as the language model's mapping goes from $\Bbb N^{1024} \to \Bbb N$ to $\Bbb N^{128} \to \Bbb N$ (and becomes closer to a bijective and thus invertible map) the masked mixer's relative training efficiency increases compared to the transformer.

    \subsection{Masked mixers are more efficient language learners than early transformer implementations}

    In some sense it is unsurprising that current transformer implementations are somewhat more efficient learners for large context windows than masked mixers because transformers have seen a number of architectural improvements and compute optimizations since their introduction. It is useful therefore to compare the learning efficiencies of early transformer implementations of all-next-token causal language modeling to masked mixers with the same training method, as these can be thought of as being similar in their developmental stage. We chose a model that was introduced before improvements such as Rotary Positional Encoding \citep{su2023roformerenhancedtransformerrotary} and Flash Attention 2 \citep{dao2023flashattention2fasterattentionbetter} and sophisticated parameter initialization techniques to name a few.

\begin{figure}[h]
        \centering
        \includegraphics[width=0.95\textwidth]{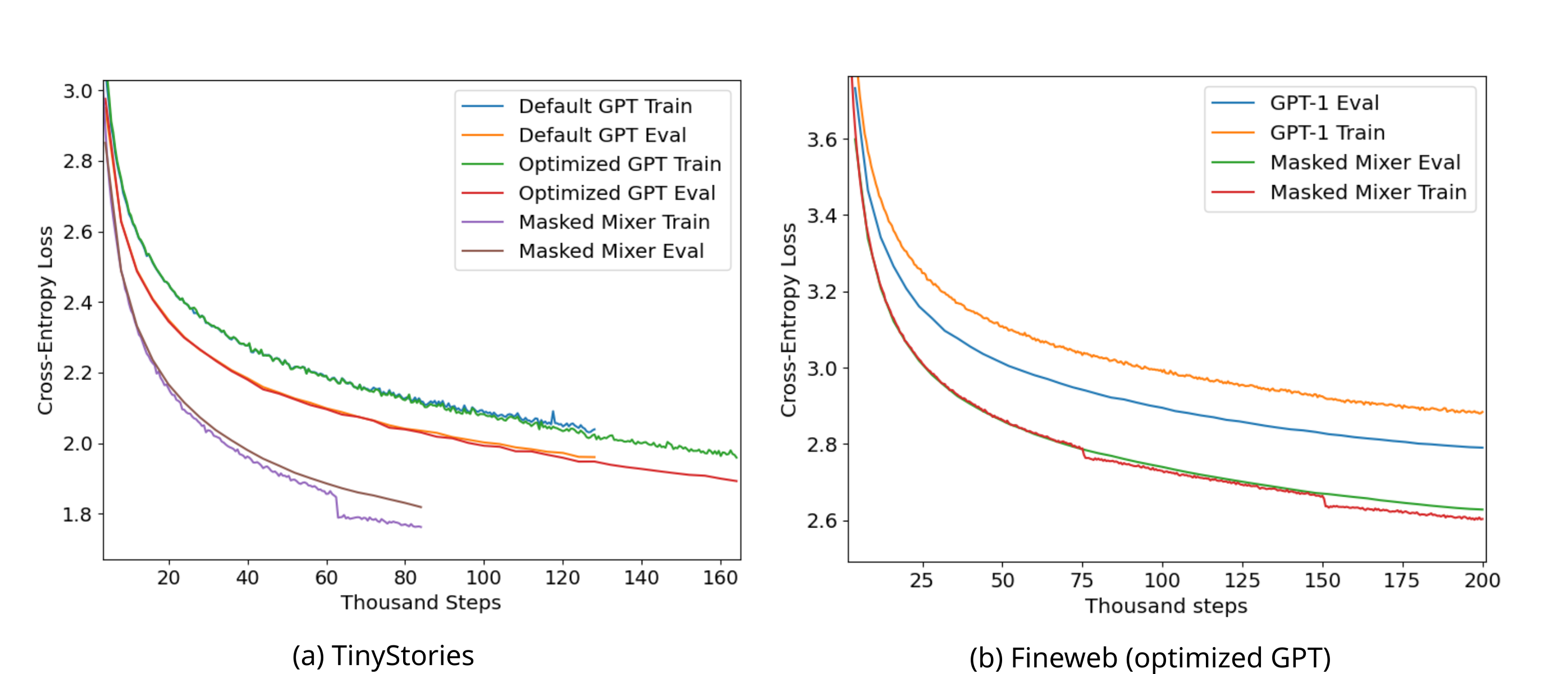}
        \caption{GPT-1 and masked mixer training efficiencies for $n_{ctx}=512$.}
        \label{fig10}
    \end{figure}

    Specifically we test the Huggingface implementation of GPT-1 \citep{radford2018improving}, a CLM-style transformer model which was introduced shortly after the original transformer architecture. We find that GPT lags well behind the masked mixer's training and validation loss values both for default as well as learning rate, batch size, and head number-optimized versions of the transformer, for both small and large datasets (Figure \ref{fig10}, Table \ref{tables6}).

\section{Alignment Between Model and Task Invertibility Predicts Training Efficiency}

    \subsection{Neither dataset nor task stochasticity explain mixer versus transformer performance differences}

    One explanation of the decrease in relative training efficiency of masked mixers to transformers as the context window increases is that transformers are better able to filter out noise via attention, where this noise could exist in the dataset itself or else in the language modeling task. We first investigated whether the gaps between transformer and masked mixer causal language model training efficiency were due to stochasticity inherent in the dataset, by which we mean the somewhat random structure inherent in natural languages. For example, observe that many equally valid next words could be used for most English sentences such that any one of these could be picked at random, as nouns and verbs and adjectives are often free interchanged and grammars are relatively inconsistent between passages.
    
    One straightforward way to test this is by observing CLM losses for each model type during training on a dataset that has a higher proportion of formal language to natural language. We chose to train on mathematical text, the FineMath 4+ dataset with approximately 10 billion tokens \citep{allal2025smollm2smolgoesbig}, as an example of such a dataset. After CLM training we find that lower cross-entropy loss is achieved as expected for a more deterministic dataset, but that once again transformers are more efficient for longer-context but not shorter-context training (Figure \ref{figs6}), supporting the hypothesis that inherent dataset stochasticity does not contribute to the transformer-mixer training efficiency gap.

    We next addressed question of whether the masked mixer and transformer training efficiency gaps were caused by the language modeling task itself, in this case using all left tokens to predict one next token. We did this by modifying the language modeling task to a two-sided prediction, rather than left-to-right prediction that is usually performed during CLM, while training on the Fineweb. Bidirectional prediction may be thought of as being more or less equivalent to the masked token prediction performed by encoding models such as BERT \citep{devlin2019bertpretrainingdeepbidirectional} except that only one token is `masked' per input. 

    Rather than use a masked token prediction method, however, we note that the training for this objective function is inherently inefficient: only a small percentage (no more that 15\% for BERT) of tokens are masked for a given sequence such that most tokens are not trained during each forward pass and merely provide context for the masked tokens. We introduce a method by which every input token is trained upon in one pass, which allows for $1/0.15 = 6.67$ times the training throughput as a standard masked language modeling approach.

    The key to this method is to note that one can use the all-next-token approach of causal language modeling in both forward and reverse directions as long as there is no more than one linear combination of forward and reverse modules per model. This is because, without loss of generality, information for a `next' token can travel between $t_{n+1}$ and $t_{n-1}$ and thence to $t_n$ if there are more than one linear combinations of forward and reverse modules (see Figure \ref{figs8} for a depiction of this process in the case of the masked mixer). Thus we split our bidirectional models into two parts, each which perform next token prediction but one that does so in the forward and one in the reverse direction.

\begin{figure}[h]
        \centering
        \includegraphics[width=0.99\textwidth]{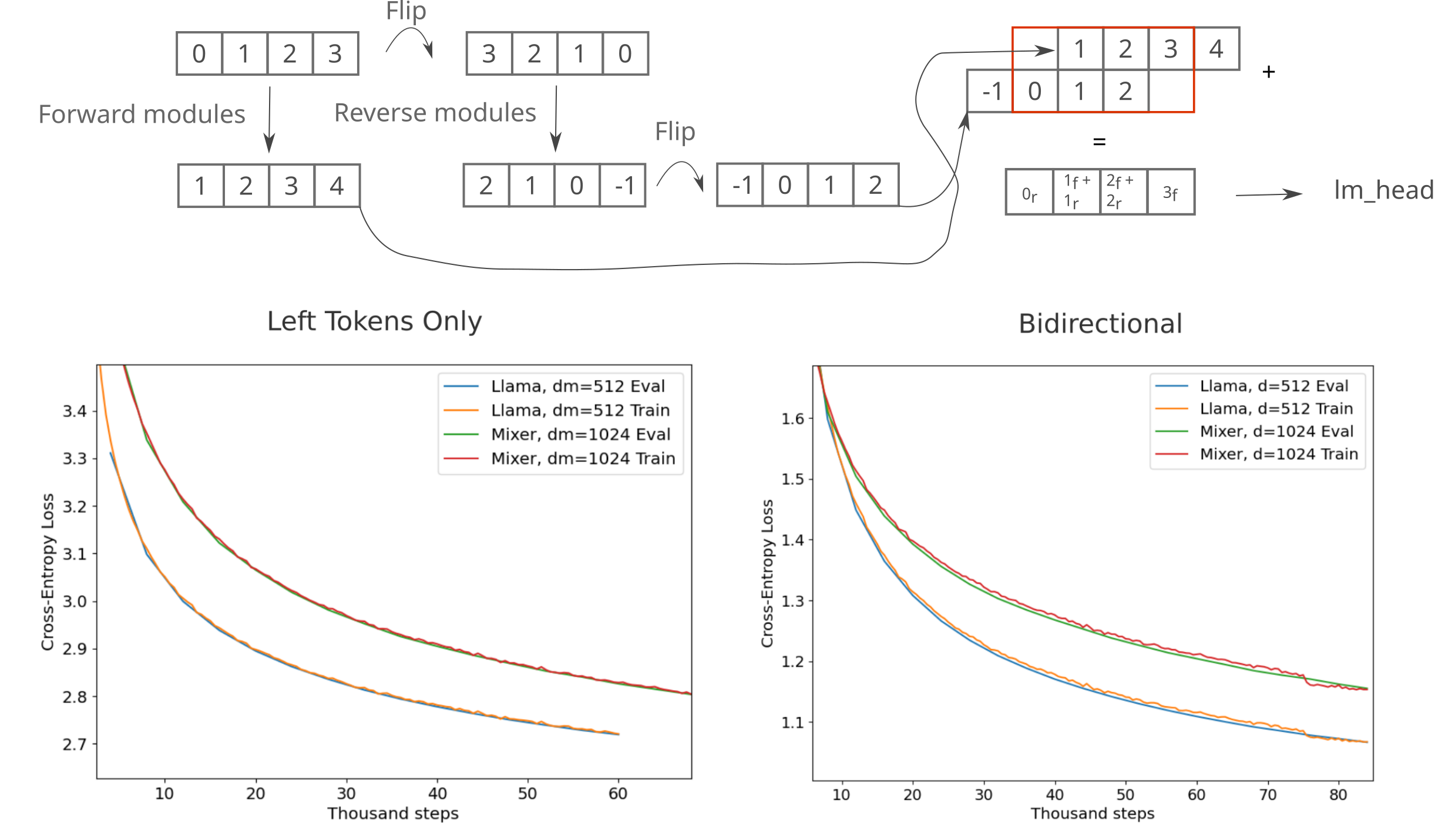}
        \caption{Top: bidirectional modeling approach with respect to token indices. Bottom: Masked Mixer and Transformer FineWeb loss during training.}
        \label{fig11}
    \end{figure}

    As shown in Figure \ref{fig11}, there is an almost identical loss gap between masked mixer and transformer for bidirectional training compared to left-token CLM training, providing evidence for the idea that task stochasticity does not explain the relative performance of each model type. 

    \subsection{Multi-token and many-token prediction efficiencies}

    After observing that masked mixers perform worse relative to transformers as $n$ increases during CLM training in which models learn $\Bbb N^n \to \Bbb N^1$ mappings, we investigated the relative training efficiencies of these models on language modeling tasks described as multifunctions of $\Bbb N^n \to \Bbb N^m, m>1$ and more specifically we start by investigating a language that may be described as a mapping $\Bbb N^{512} \to \Bbb N^2$.

\begin{figure}[h]
        \centering
        \includegraphics[width=0.99\textwidth]{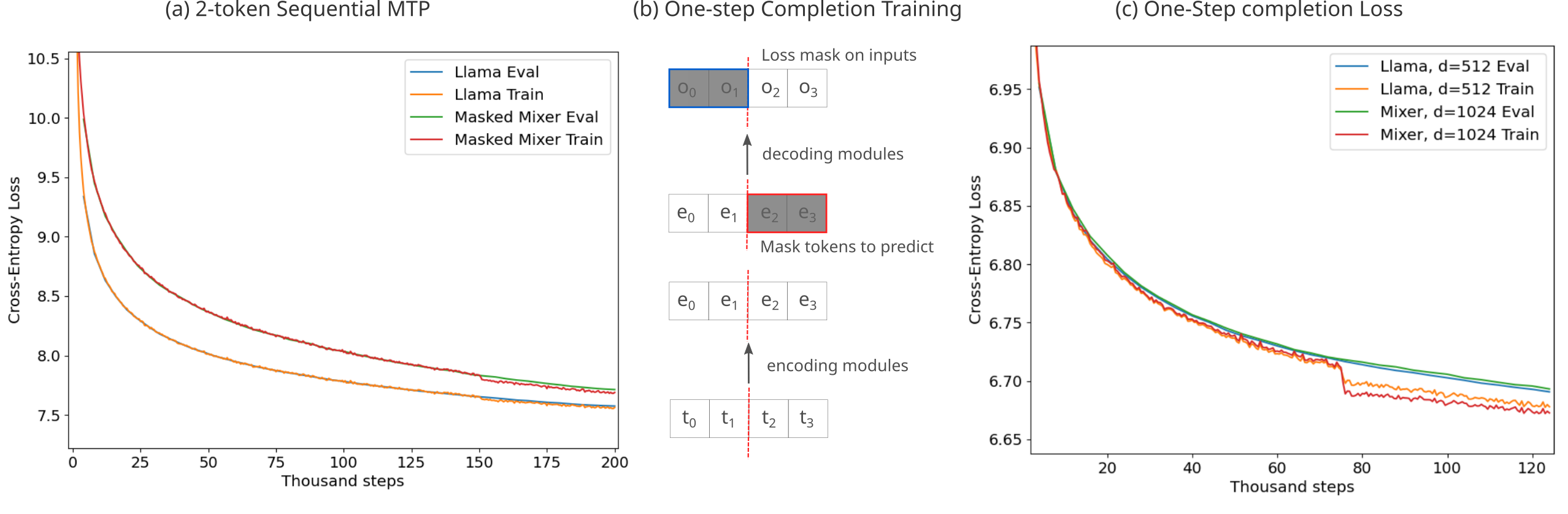}
        \caption{FineWeb training curves for sequential multi-token prediction and parallel many-token prediction. Masked Mixers are $d_m=1024, n_l=16$ and transformers are llama-style $d_m=512, n_l=16$ with $n_{ctx}=512$ for both experiments.}
        \label{fig15}
    \end{figure}

    The assumptions behind training a language model to predict one next token at a time is that this is an easier task than predicting many next tokens at once, and that simply being able to predict each next token allows for effective inference. The latter implies that a greedy strategy during training is best, but the ability of inference methods such as beam search to outperform greedy inference suggests that the training a model to predict multiple `next' tokens may be beneficial. This has recently been referred under various names such as ‘multiple token prediction’ or ‘non-myopic pretraining’, and there are many method by which one could implement this idea ranging from a multi-headed approach where each head predicts one token \citep{gloeckle2024betterfasterlarge} to the more direct approach of performing $m$ forward passes and add the corresponding loss values for each of $m$ tokens predicted, performing label shifting with a shift position size corresponding to the number of forward passes similar to the method used in \citep{deepseekai2025deepseekv3technicalreport}. We take the latter approach and perform two forward passes through the model for two-token prediction. As shown in Figure \ref{fig15} (a), the loss gap between masked mixer and transformer is slightly smaller for two-token prediction compared to one-token prediction at 1.8\% and 2.0\%, respectively (refer to Figure \ref{fig13}).

    We also investigated the efficiencies of language models trained to predict many next tokens in one forward pass instead of only one such that each token prediction is $\Bbb N^{n/2} \to \Bbb N^1$ with all token predictions being $\Bbb N^{n/2} \to \Bbb N^{n/2}$ which was achieved by the architecture shown in Figure \ref{fig15} (b). Somewhat arbitrarily choosing to generate half of the text (meaning a context window of 512 we use the first 256 tokens as inputs and perform loss computation on the next 256 tokens), we find that a masked mixer and transformer yield nearly identical Fineweb training efficiencies Figure \ref{fig15} (c).

    \subsection{Masked mixers are far better autoencoders than transformers}
        \label{autoencoding}

\begin{figure}[h]
        \centering
        \includegraphics[width=0.99\textwidth]{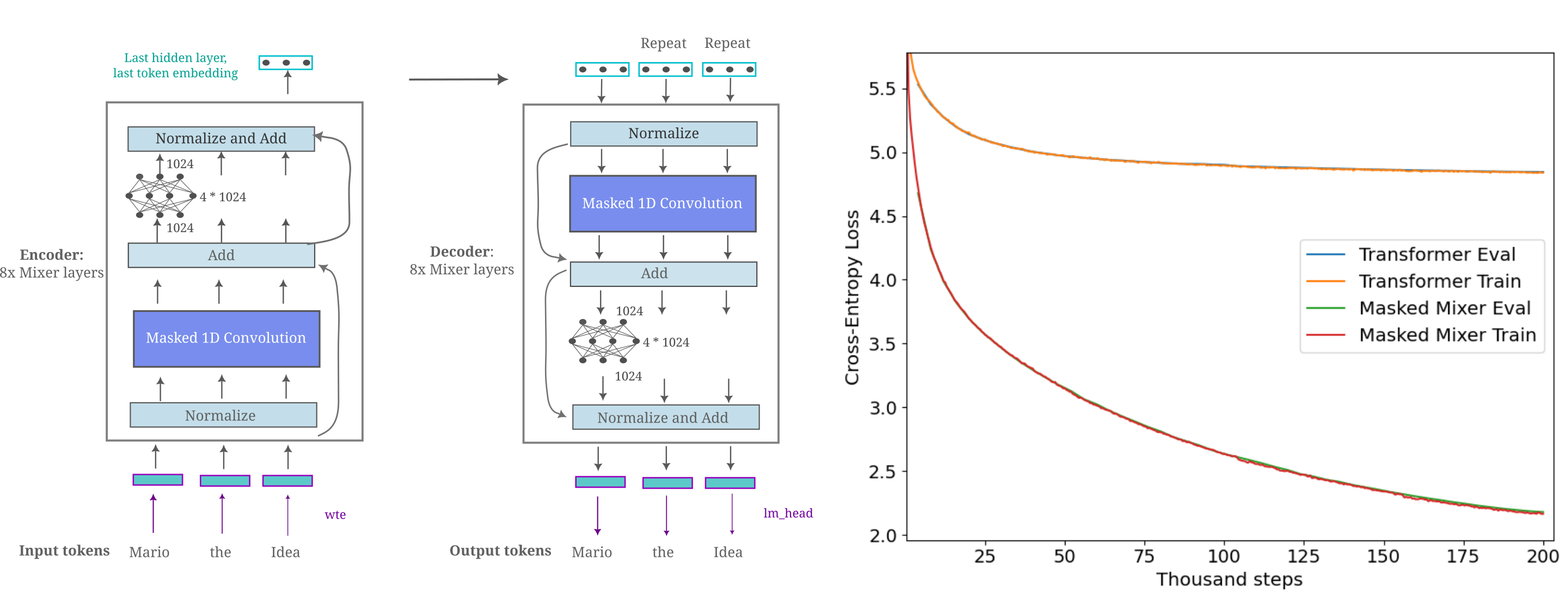}
        \caption{Left: Mixer Autoencoder architecture. Right: Autoencoder loss on FineMath 4+.}
        \label{fig12}
    \end{figure}

     At the other extreme of the $\Bbb N^n \to \Bbb N^1$ mapping that is performed during next token prediction, we test masked mixer and transformer training efficiencies on a task that requires an $\Bbb N^n \to \Bbb N^n$ (i.e. invertible) transformation: a non-trivial (with respect to inter-token information flow) autoencoding in which an encoder takes the input and compresses it to one hidden layer, and the decoder takes this hidden layer and attempts to regenerate the input tokens. Such an autoencoder may be implemented by using a encoder-decoder model where a standard masked mixer acts as the encoder, the last hidden state of the last token is taken as the embedding, and a decoder composed of a stack of masked mixer modules transforms this embedding (repeated for each input word) into a sequence to match the input as shown in Figure \ref{fig12}. We implement an analogous architecture using a modified Llama-style transformer without word-token embedding layers or language modeling heads for encoder and decoder modules.
     
     This mixer autoencoder reaches a cross-entropy loss on FineMath comparable to non-optimized transformers and mixers (albeit with around twice the number of training steps required) and very little or no overfitting, indicating that the embedding created by the model's encoder provides generalizable information on the input to the decoder, and in particular enough information for the decoder to reconstruct the entire input sequence with accuracies approaching all-next-token causal language models trained using identical compute (Figure \ref{fig12}). In contrast, transformers are far less efficient for this autoencoding and are unable to predict input tokens with anywhere near the accuracy of causal language modeling-trained transformers or masked mixers or mixer autoencoders (Figure \ref{fig12}). We observe this to be the case for TinyStories as well, and confirm that these differences are not caused by differences in $d_m$ between transformer and masked mixer (Figure \ref{figs10}).

\section{Masked Mixers are Better for Retrieval}

    One of the most frequently encountered language tasks today is retrieval, a task which usually involves finding one or more matches (the `target') for a given language segment (the `query') given a set of potential matches. One of the most common uses for this is in retrieval-augmented generative search, whereby a search phrase is `read' by a language model and the resulting embedding from that phrase is then used to match one or more segments of a corpus of text, which is in turn fed to a generative model along with the original search phrase \citep{lewis_retrieval}. 

    As is the case for language generation, nearly every retrieval model used today is based on the transformer architecture. The results in previous sections of this work lead to the hypothesis that attention is not well-suited for the task of retrieval because these transformations are biased towards non-invertibility, such that a mapping with this transformation becomes many-to-one (Figure \ref{fig3}). For predicting a single next token it appears that this mapping is not detrimental and may even be beneficial, as masked mixers learn to reduce the information passing to deep layers during causal language model training.
    
    On the other hand, matching a sequences of tokens to another sequence of tokens may be thought of as approximations to a bijective function which would not be expected to benefit from information reduction or non-invertibility. Masked mixers have been shown to be effectively invertible for small inputs and in that respect retain much more input information than mixers, suggesting that these models are better suited to retrieval than models with attention assuming that retrieval requires most input information.  In this section we test this idea experimentally two ways, and find that indeed last-hidden-layer embeddings from masked mixers are far superior for retrieval tasks compared to the same embeddings from transformers in both settings.
    
    \subsection{Synthetic dataset generation and embedding}

    To investigate retrieval abilities of transformers and masked mixers on in-distribution data while maintaining flexibility as to the number of samples included per retrieval, we generated a synthetic summary sentence for the first 200k samples (truncated to 512 tokens maximum) in the FineWeb dataset using Llama-3 (8b) Instruct \citep{dubey2024llama} by prompting that model to read each 512-token-length text segment and give a one-sentence summary of that segment. We found that this relatively small model was more than capable of accurately and concisely summarizing most target segments. We repeated this procedure with the FineMath 4+ dataset, for either 200k or 400k samples, and changed the model prompt to make the summary query more concise (and more difficult to match).

    Each text segment and its corresponding summary were fed to a pretrained Llama-style transformer or masked mixer and the last hidden layer activations were saved as the model's embedding. We use the second-to-last token's output for the embedding because the `last' masked mixer token is untrained due to the index shift that occurs during all-next-token training of these models. This procedure only rarely would be expected to lead to input information loss, as the synthetic summary queries nearly all contain ending punctuation tokens. As masked mixers have an implicitly defined absolute positional encoding, we pretrain these models on left-padded tokens rather than the right-padded token segments used elsewhere in this work in order to be able to use the second-to-last token hidden layer for inputs of all sizes. Training with left-padded tokens leads to some decrease in CLM training efficiency as no weight masking is performed on these inputs. We experimentally determined that left versus right padding during pretraining yields virtually no difference for subsequent retrieval training for transformers, as long as the retrieval training itself proceeds with left padding. For consistency, we pretrained transformers used for retrieval with left-padded inputs as well.

    \subsection{Indirect retrieval model architecture and training approach}

    We first investigated the ability of masked mixers and transformers to perform retrieval after next-token causal language model pretraining where the model performing the embedding remains unchanged during retrieval training. This allows us to address the question of how effective each model's pretrained embeddings are for retrieval while us to reference these values to existing invertibility measurements (Figure \ref{figs5}). We note that there are large training throughput gains with this method as well as retrieval training typically proceeds by modifying the parameters of the embedding model directly \citep{wang2024improving} which is efficient for inference but in some sense is inefficient for training because one forward pass is required for each input for each batch of matching and non-matching sequences.  If one first embeds each input using a trained generative model and then trains a separate model on these embeddings then a single forward pass is required for comparison of all inputs in each forward pass of an input size $c = n_{ctx}$ where $c$ is the number of embeddings. 

    For our retrieval model we chose to implement a mixer with bidirectional convolutions rather than masked ones, similar to MLP mixers but with only one convolutional layer rather than two between sequence elements, as transformers were found to be very poor retrieval models even for small comparison contexts (Table \ref{tables8}). Word -token embedding layers are removed (as the inputs are embeddings already) as are language heads, and instead each last hidden layer has a $d_{model} \to 1$ transformation where the unitary output corresponds to the likelihood (after Softmax transformation) of that input embedding being the correct match. Softmax transformation serves to stabilize the retrieval model's loss, as increasing all logit values does not decrease the cross-entropy loss, and a depiction of this architecture for masked mixers is found in Figure \ref{fig16}. 

    Arbitrarily choosing the first `token' to be the summary and all other tokens to be the potential matching target text segment embeddings, each input is then assembled by randomly sampling all target embeddings before replacing one input (at a random location) with the embedding of the matching target.  A non-optimized version of this approach for generating inputs for the retrieval model is found in Algorithm \ref{alg1}. 

\begin{figure}[htb]
  \centering
  \begin{minipage}{.85\linewidth}
    \begin{algorithm}[H]
    \caption{Retrieval training dataset sampling}
    \label{alg1}
    \begin{algorithmic}[1]
        \Require $x = O_l(\mathrm{queries}, \theta_g)$ \Comment{Query embeddings}
        \Require $y = O_l(\mathrm{targets}, \theta_g)$ \Comment{Target embeddings}
        \Require c $\gets$ len(context) \Comment{Number of samples trained per match}
        \Require vector a: len(a) = c \Comment{Vector of embedding vectors for model input}
        \Require vector q: len(q) = c, q[i] = 0 \Comment{Label vector}
        \State $N \gets |x| = |y|$
        \While{n $\gets$ 0, n++, n < N} \Comment{Loop over query/target pairs}
            \State a[0] $\gets$ x[n] \Comment{First input element is the query to match}
            \State weights: weights[0, 1, ... N] $\gets$ 1
            \State weights[n] $\gets$ 0
            \State r = multinomial(weights, c-1) \Comment{Random sample of all input indices except matching}
            \State a[1:c] $\gets$ y[r] \Comment{Replace elements of a with lookup of embeddings}
            \State m = randint(1, c) \Comment{Random index to place matching target embedding}
            \State a[m] $\gets$ y[n] \Comment{Matching target embedding placed}
            \State q[m] $\gets$ n \Comment{One-hot label at position m}
            \State yield a, q, m
        \EndWhile
    \end{algorithmic}
    \end{algorithm}
    \end{minipage}
\end{figure}

    The outputs after Softmax transformation are compared to the true distribution via standard cross-entropy loss, which is then back-propagated through the retrieval model but not further to the generative model or input embedding. Note that the logits are not Softmax-transformed during inference as this operation has no effect on top-k choice without noise introduction. We use a relatively modest batch size (512). We make use of the Pytorch Cross-Entropy Loss function's compatibility with class probabilities for this model. In contrast to the rest of this work, the efficiency of an embedding for retrieval modeling is not measured in loss per compute amount or even loss per number of samples seen but instead minimum evaluation cross-entropy loss over a pre-determined.
    
    The evaluation set cross-entropy loss after indirect retrieval training for a $d_m=512$ masked mixer and two Llama-style transformers pretrained on the FineWeb dataset are shown in Figure \ref{fig16} and give evidence for the hypothesis that masked mixer embeddings are superior for retrieval training compared to those from transformers. Notably, training fails (with no training loss minimization) when using embeddings from models that were not pretrained, or if the retrieval model is implemented with a transformer architecture instead of a mixer.

    \begin{figure}
      \begin{minipage}[b]{.42\linewidth}
        \centering
        \includegraphics[width=0.92\linewidth]{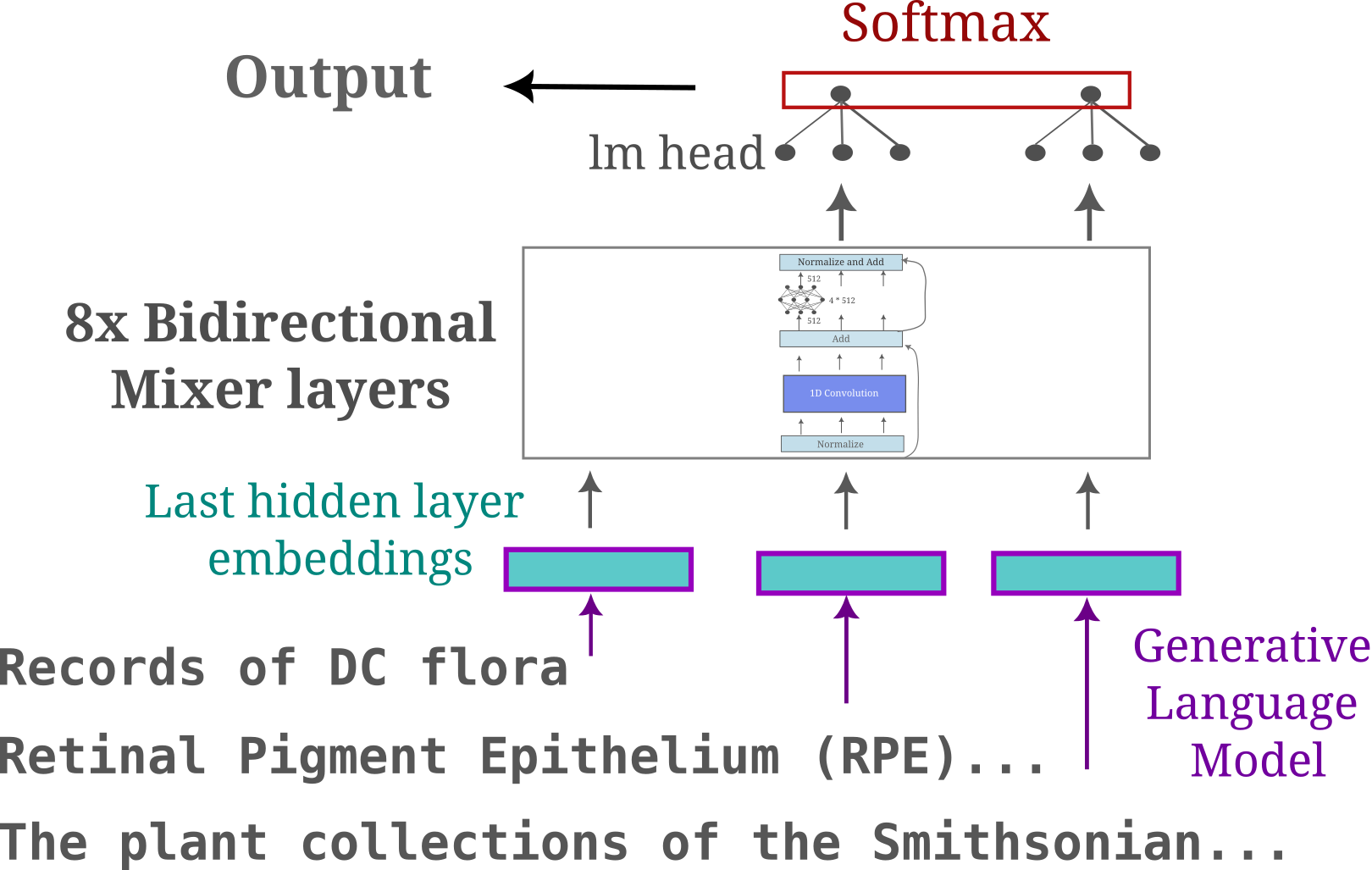}
      \end{minipage}\hfill
      \begin{minipage}[b]{.58\linewidth}
        \centering
        \renewcommand{\arraystretch}{1.2}
        \begin{tabular}{||l c c c c c ||} 
         \hline
          Model & $n=32$ & 128 & 256 & 512 & 1024 \\ [0.5ex] 
         \hline\hline
          Masked Mixer & 0.28 & 0.58  & 0.78 & 0.88 & 1.20 \\
         \hline
          Transformer, h=4 & 0.55 & 1.24 & 1.76 & 1.84 & 6.93  \\ 
         \hline
          Transformer, h=32 & 0.61 & 1.19 & 1.59 & 1.88 & 6.93 \\ 
         \hline
        \end{tabular}
        \vspace{15px}
      \end{minipage}
      \caption{Retrieval mixer architecture and FineWeb retrieval evaluation cross-entropy loss ($n$ samples per retrieval model forward pass, sampled without replacement).}
      \label{fig16}
    \end{figure}

    \subsection{Masked mixers are more suitable for cosine similarity InfoNCE -trained retrieval}
    \label{infonce}

    The two major downsides to the indirect retrieval model training approach is that there is no guarantee that the embedding model is capable of capturing the information necessary to perform retrieval as the embedding model was trained for an entirely different task (causal language modeling in this case), and that inference requires much more computation assuming that the retrieval model is composed of more than one matrix multiplication.  Empirically the first is observed to be the case for FineWeb as the retrieval model tends to overfit the retrieval dataset. It is therefore pertinent to observe the ability of the transformer and masked mixer to be trained to form embedding suitable for retrieval and one particularly promising way to do this is to perform noise contrastive training on a causal language-pretrained model.
    
     We implement a temperature-scaled cosine distance-based InfoNCE loss where $q^+$ indicates the query embedding, $\tau=0.02$ the temperature hyperparameter set to 0.02, $d^+$ the matching text embedding and $d_i$ randomly sampled non-matching text embeddings as shown in Equation (\ref{eq9}), after \citep{wang2024improving}.

    \begin{equation}
    \Bbb{L} = - \log \frac{f(q^+, d^+)}{f(q^+, d^+) + \sum_i f(q^+, d_i)}
    \label{eq9}
    \end{equation}
    
    For (\ref{eq9}) we use the model's last hidden layer of the second-to-last token as the embedding for cosine similarity computation shown in Equation (\ref{eq10}) instead of appending an extra token to each sequence, and sample with a context window size of 512. We include no hard negatives and sample one query with one matching and thirty non-matching text segments, and process four batches per optimizer update (one per GPU). We sample each batch using an analogous method to Algorithm \ref{alg1} except that token sequences rather than embeddings are retrieved, and each input is assembled in the batch dimension such that a batch size of 32 has 32 separate text sequences.

    \begin{equation}
    f(a, b) = \exp \left(\frac{1}{\tau} \cos(O(a, \theta), O(b, \theta)) \right)
    \label{eq10}
    \end{equation}

    We use a standard approach to retrieval inference via the cosine similarity identity defined in Equation (\ref{eq11}) which we implement in an efficient batched form by first normalizing last hidden layer activations of the query $x$ and n targets $y_n$ such that $||x|| = ||y_n|| = 1$ before performing vector multiplication on the query output $x$ and row-concatenated target vectors $Y$ such that the output is $z = xY^T$, where the top-1 value of the output vector $z$ is found in the batch dimension.

    \begin{equation}
    z_n = \cos (x, y_n) = \frac{x \cdot y_n}{||x|| \; ||y_n||}
    \label{eq11}
    \end{equation}

    We first benchmarked a state-of-the-art (when this work was commenced) large model pretrained on trillions of tokens and postrained via cosine similarity InfoNCE for retrieval (among other tasks), e5 Mistral Instruct \citep{wang2024improving}. When we applied to the synthetic Fineweb retrieval dataset introduced in the last section we observed that top-1 evaluation accuracy for 32- and 64-sample retrieval of around 100\% for this model (compared to 92.8\% and 87.7\% respectively for the masked mixer as trained using the method above) suggesting that this dataset is not particularly challenging for a strong model like e5 Mistral Instruct. We therefore generated a new synthetic dataset in which the same Llama model was instructed to generate shorter queries for FineMath 4+ text segments. This can be viewed as being a more difficult retrieval task because the dataset itself is restricted in subject and also because the queries are themselves less shorter and descriptive than they were for the FineWeb. 

    The top-1 retrieval accuracy of each model with the FineMath 4+ evaluation Cross-Entropy Loss for models trained on this dataset are shown in Table \ref{table9}. For both top-1 retrieval accuracy and loss values, measurements were performed on evaluation datasets held out during each respective training, with the pretraining corpus including both training and evaluation target text segments used for retrieval and none of the queries. All models except the already-trained e5 Mistral Instruct were trained using n=180k retrieval samples (one sample being a query with its matching target text) except where noted, and evaluated on a hold-out dataset of 20k samples. Note that e5 Mistral Instruct uses a different tokenizer such that its cross-entropy loss is not comparable, and that due to the masked mixer's far fewer inter-token parameters the transformer in the first line of Table \ref{table9} was pretrained using roughly the same compute as the $n_l=32$ and $d_m=1024$ masked mixers and around double the compute as the masked mixer in the second line of that table.
    
    \begin{center}
    \begin{table}
    \begin{center}
    \renewcommand{\arraystretch}{1.2}
    \begin{tabular}{||l c c||} 
     \hline
      Model & Top-1 Retrieval Accuracy (\%) & Pretraining Loss \\ [0.5ex] 
     \hline\hline
      Transformer & 66.2 & \textbf{1.39} \\ 
     \hline
      Masked Mixer & 81.2 & 1.79 \\ 
     \hline
      Masked Mixer $n_l=32$ & 84.6 & 1.65 \\ 
     \hline
      Masked Mixer $d_m=1024$ & 86.0 & 1.53 \\ 
     \hline
      Transformer s=380k & 95.0 & \textbf{1.39} \\
      \hline
      e5 Mistral Instruct $d_m=4096$ & 95.1 & -- \\ 
     \hline
      Masked Mixer $d_m=1024$, s=380k & \textbf{98.2} & 1.53 \\ 
     \hline
    \end{tabular}
    \end{center}
    \vspace{0.1cm}
    \caption{FineMath evaluation retrieval top-1 accuracy with n=32 sample size per retrieval, postrained on s=180k synthetic retrieval samples with $n_l=16$ and $d_m=512$ except where denoted.}
    \label{table9}
    \end{table}
    \end{center}

    For practical query retrieval tasks such as large scale RAG, one usually wants to perform retrieval using sample sizes of hundreds or thousands of targets. This is a major drawback for benchmark datasets which compare queries to only a few targets (for example as is typical for datasets like MSMarco \citep{bajaj2018msmarcohumangenerated}) because few-sample retrieval performance has no guarantee of extrapolating to many-sample retrieval. It is therefore important to measure many-sample retrieval performance directly, and the these results are shown in Tables \ref{table10} and \ref{table11}. Most notably, the transformer model's near-parity performance to the much larger e5 Mistral Instruct on small-context inputs vanishes as the sample size increases such that it becomes far worse for large-context retrieval relative to e5 Mistral, whereas the masked mixer (trained on identical data) exhibits uniformly superior performance to the other models across all sample sizes.

    \begin{center}
    \begin{table}[H]
    \begin{center}
    \renewcommand{\arraystretch}{1.2}
    \begin{tabular}{||l c c c c c c c c c||} 
     \hline
      Model & $n=32$ & 64 & 128 & 256 & 512 & 1024 & 2048 & 4096 & 8192 \\ [0.5ex] 
     \hline\hline
      Transformer & 95.0 & 92.3 & 88.9 & 84.5 & 79.1 & 72.6 & 65.7 & 57.6 & 50.4 \\
      \hline
      Transformer Autoencoder & 94.4 & 92.0 & 88.4 & 84.3 & 79.1 & 73.1 & 67.0 & 60.6 & 53.3 \\
      \hline
      Mixer Autoencoder & 97.1 & 95.4 & 93.1 & 90.0 & 86.3 & 82.1 & 76.7 & 70.7 & 64.5 \\
      \hline
      Mixer Autoencoder, pretrained 500k & 96.9 & 95.5 & 93.1 & 90.0 & 86.6 & 82.0 & 76.8 & 71.5 & 65.0 \\
     \hline
      e5 Mistral Instruct & 95.1 & 93.3 & 90.9 & 88.5 & 85.6 & 82.2 & 78.4 & 73.5 & 68.8  \\ 
     \hline
      Masked Mixer & \textbf{98.2} & 97.2 & \textbf{95.8} & 93.4 & 90.2 & 86.5 & 81.7 & 76.6 & 70.6 \\ 
     \hline
      Masked Mixer, pretrained 500k & \textbf{98.2} & \textbf{97.3} & \textbf{95.8} & \textbf{93.5} & \textbf{90.7} & \textbf{86.8} & \textbf{82.4} & \textbf{77.3} & \textbf{71.7} \\
     \hline
    \end{tabular}   
    \end{center}
    \vspace{0.1cm}
    \caption{FineMath 4+ evaluation retrieval top-1 accuracy (\%) for various sample sizes $n$, transformer and mixers were pretrained for 200k steps except where noted, and post-trained on s=380k synthetic retrieval samples}
    \label{table11}
    \end{table}
    \end{center}

    As has been observed for transformers, increasing the masked mixer's pretraining compute increases the resulting model's retrieval performance (Table \ref{table11}), and pretraining on an I.I.D. dataset to the retrieval task yields increased retrieval performance than pretraining on a dataset with a mix of in- and out of- distribution data such as that in the FineWeb (Table \ref{tables8}). These effects are dataset-dependent, however, as we find accurate FineWeb retrieval using models pre- and post-trained on FineMath 4+ (Table \ref{tables12}). For models trained on the FineWeb and evaluated on FineMath we find that the masked mixer is once again more accurate than the transformer, especially for large sample sizes (Table \ref{tables12}).

    From the large increase in accuracy upon dataset size doubling, it may be wondered whether doubling the dataset again would allows the transformer to match the performance of the masked mixer with a smaller dataset. When this is done, we find that the transformer gains little accuracy and is not near matching the masked mixer's accuracy, particularly for large-sample retrieval (Table \ref{table12}). A preliminary scaling study of top-1 accuracies achieved for transformers and masked mixers postrained with various dataset sizes (for retrieval with $s=8192$) is shown in Figure \ref{figs11}, and demonstrates that one would require many orders of magnitude more retrieval training data for the transformer to match the mixer's accuracy.

    \begin{center}
    \begin{table}[H]
    \begin{center}
    \renewcommand{\arraystretch}{1.2}
    \begin{tabular}{||l c c c c c c c c c||} 
     \hline
      Model & $n=32$ & 64 & 128 & 256 & 512 & 1024 & 2048 & 4096 & 8192 \\ [0.5ex]
     \hline\hline
      Transformer & 95.8 & 93.3 & 90.2 & 86.0 & 80.8 & 74.4 & 67.6 & 60.5 & 52.6 \\
      \hline
      Masked Mixer & 98.6 & 97.8 & 96.5 & 94.5 & 91.7 & 88.5 & 83.8 & 79.0 & 73.7 \\
     \hline
    \end{tabular}   
    \end{center}
    \vspace{0.1cm}
    \caption{FineMath 4+ evaluation retrieval top-1 accuracy (\%) for various sample sizes $n$, post-trained on s=780k synthetic retrieval samples}
    \label{table12}
    \end{table}
    \end{center}

    \subsection{Autoencoders yield effective pretrained embeddings for InfoNCE}

    We also investigated the ability of a mixer pretrained on the FineWeb dataset using the autoencoding training presented in Section \ref{autoencoding}, modifying the InfoNCE training protocol accordingly to use the last token of the model's encoder rather than second-to-last token's embedding, and discarding the decoder. After InfoNCE retrieval training on FineMath, we find that this model achieves a higher 32-sample top-1 FineMath 4+ retrieval accuracy than the masked mixer pretrained via next token prediction (Table \ref{tables8}), despite the encoder containing half the number of layers as the CLM-pretrained masked mixer, and is even more accurate than the transformer trained on FineMath itself. Pretraining the mixer autoencoder on FineMath yields better accuracy than the transformer applied to the same dataset, but less accuracy than the CLM-pretrained masked mixer (Table \ref{table11}).  
    
    Masked mixers and transformers without pretraining are virtually untrainable using the cosine similarity-based InfoNCE method, as these models fail to minimize the objective function value to any significant degree in our experiments. Curiously, transformers pretrained to autoencode using this method yield effective retrieval models after InfoNCE training and indeed are more accurate large-sample retrievers than CLM-pretrained transformers, despite the very poor performance of the transformer with respect to minimization of the autoencoding objective function (Table \ref{table11}, Figure \ref{fig12}). In general it seems that minimization of the autoencoding objective function does not yield benefits for retrieval beyond a certain point, as continuing the mixer autoencoder pretraining to 500 thousands steps yields a much lower evaluation loss (CEL of 1.38) compared to the standard 200k steps (CEL of 2.18) but very little change in retrieval performance post-InfoNCE (Table \ref{table11}).
    
    We conclude that autoencoding provides a promising new direction for pretraining for retrieval models as this pretraining method appears to be far less sensitive to the amount of compute used, although more work needs to be done to more fully determine the strengths and weaknesses of this approach as compared to causal language modeling-based pretraining.
    
\section{Limitations Of This Work}

    This study is limited in scale: experiments were performed using a relatively small datasets (none larger than 10 billion tokens) and compute (four V100 GPUs) and model sizes (tens or hundreds of millions of parameters). Although there are reasons to suspect that the causal language modeling results here would translate to much larger datasets and compute, this has yet to be investigated as training runs were performed in 30h or less on 4x V100s (for a maximum theoretical limit of $5 * 10^{19}$ Tensor FLOPs), which is a relatively small amount by modern standards. Another limitation inherent in this work is that compute efficiency is measured in a concrete hardware-specific fashion. We chose this route as it is immediately applicable to the primary goal of language model training (ie getting the lowest possible loss in the least amount of time), and one can recover the FLOPS per experiment using available data on each compute node. 
    
    A downside to our wall clock and hardware-specific performance measurements is that our GPUs are not typically used for language model training in industrial settings today, where H100 or A100s are favored. Newer GPUs are specially tailored for transformer inference and training and thus our comparisons between transformer and mixer architectures would likely be somewhat different if performed on these newer GPUs. H100s in particular have dedicated transformer engines \citep{nvidiah100}, which would be expected to provide speedups relative to other architectures. Likewise, our approach used one set of software implementations (Pytorch, Huggingface Transformers, Accelerate etc.) and our conclusions with respect to training efficiencies are limited by the choices made in those libraries.

    Our input representation methods yield results that are consistent with other invertibility investigations (autoencoding performance for example) but are by no means comprehensive. It is entirely possible that a different input representation method would yield different results, however.
    
\section{Discussion}

    \subsection{Input representation accuracy and training efficiency: what good is invertibility for deep learning?}

    This work has explored the use of input representation accuracy and global invertibility as a guide to improving deep learning model architectures for language generation and retrieval, and has found that global invertibility is indeed very useful in estimating performance capabilities. This is not at all a surprising result when viewed from the theory of numerical analysis, as invertibility is usually the second-most important facet of any numerical framework (after linearity). It has also become is evident that there is no simple relationship between representation accuracy and either generation or retrieval or else very small masked mixers would outperform transformers, and untrained mixers would for effective embeddings for retrieval. It is evident that other considerations (for example, the set of possible transformations a model is capable) are important in varying degrees as well.

    One might wonder if accurate input representation would contribute to a model's ability to overfit the inputs it was trained on. In this work we found that indeed the gap between training and test loss is slightly higher for masked mixers compared to transformers and in that sense the answer is yes. But we have yet to see any examples of severe overfitting of a language dataset of significant size for any masked mixer or transformer, which provides support for the ideas that memorization as it is currently observed is often a property of the dataset itself rather than a model's architecture and that even very large models capable of memorizing large datasets are intrinsically biased against doing so when training using gradient descent due to a few characteristics of high-dimensional space \citep{badger2022deeplearninggeneralizes}. 

    \subsection{Semi-recurrent and feedforward architectures}

    Transformers blend features of recurrent neural networks and pure feedforward models: during training they require only one forward and backward pass per input sequence to generate useful gradients using the loss from each output sequence element (like feedforward models) but during inference they may be used for language generation of an indeterminate number of tokens (assuming that the positional encoding method used is sufficiently flexible).  The main downside is that these models are effectively $O(n^2d)$ in both time and space complexity during training because for each next token for an input of $n$ tokens, the key and value projections must be re-computed.  Masked mixers as implemented are pure feedforward models in the sense that these models have a fixed output size for both training and inference. 

    We can also comment on similarities and differences between the masked mixer and other language models replacing attention with specialized convolutions such as the Monarch mixer \citep{fu2023monarchmixersimplesubquadratic} or Hyena \citep{poli2023hyenahierarchylargerconvolutional} architectures, which are motivated primarily by an effort to reduce the $O(n^2d)$ computational complexity inherent in the transformer architecture.  Our work has an entirely different motivation, namely increased representational efficiency rather than lower bounds for computational scaling, and thus focuses on training efficiency at the expense of inference efficiency. In particular, masked mixers have higher inference complexity ($O(n^3d)$) than transformers, which can cache hidden layer values. Caching masked mixer hidden state values during inference would be useful for convolutions with restricted substructure (for example, parameterized as Toeplitz or Monarch matrices) but we leave the investigation of such models to future work.

    \subsection{How much more efficient are masked mixers compared to transformers for retrieval?}

    The results in Section \ref{infonce} indicate that masked mixers are substantially more effective retrieval models compared to transformers receiving the same training protocol, and we include results from the near-SOTA e5 Mistral instruct as a benchmark of what a very capable off-the-shelf model can do for this dataset. 
    
    That said, it is helpful to note the scale of the computational resources with which these masked mixers were trained relative to e5 Mistral. It is unknown exactly how many GPUs and for how long the Mistral 7b model was pretrained, or if this model was indeed trained from scratch without weight transfer, but interpolating this model’s performance with others for which data is available such as Llama 2 (8b) which required 184k A100 GPU-hours and Llama 3.1 (8b) which required 1.46M H100 hours suggests that the 7b Mistral model would require perhaps 500k A100 GPU-hours for pretraining, as a rough ballpark figure. Assuming an equivalence of 1.5 * V100 = A100 (which is empirically a fairly accurate value for for 16/32 bit mixed precision training) we get 750k V100 GPU-hours, which is approximately 10,000x the amount of compute that the transformer and masked mixer was pretrained with (80 GPU-hours). The e5 Mistral Instruct model was then retrieval-trained with 32 V100s over days, which is between 100x and 1,000x the compute we used here for our retrieval training process.

    Another way to measure the relative efficiency of the transformer versus the mixer for retrieval is to find what amount of training data is necessary for one model's performance to match the other's. The transformer's accuracy for $s=8192$ retrieval increases by 31.6\% and 2.2\% for two dataset doublings (200k to 400k to 800k total dataset samples) suggest that one would need to transform this dataset size by $31.6/2.2 > 14$ logarithms for linear interpolation, yielding an impossibly large number of samples necessary for the transformer to match the masked mixer's performance. If we relax this value by a large amount and instead interpolate once (such that every next dataset doubling yields $0.153$ percent accuracy increase, one requires a dataset $2^{(70.6-50.4)/0.153} \approx 2^{132} > 5 * 10^{39}$ as large as that for the masked mixer.

    Finding exactly how much more efficient the masked mixer architecture is for retrieval would require more knowledge of the scaling properties of each model type with respect to parameter number, pretraining dataset size, and post-training dataset size. We leave this work to future investigations.

    \subsection{Is attention useful for efficient language modeling?}

    The MLP mixer was originally introduced to test the importance of attention in the transformer architecture's effective modeling, and it was found that swapping attention for MLPs does not lead to a large drop in vision modeling performance but that swapping MLPs for attention does \citep{melas2021you}. In this work we have found that attention is not necessary for training an effective generative language model as well, but primarily focus on the efficiency of the training process rather than its asymptotic characteristics. 

    Creating an efficient learning process is in some ways the fundamental goal nearly every deep learning architecture: as a one-hidden-layer feedforward network is universal for all computable functions \citep{hornik1989multilayer}, if efficiency was not a concern then that is the only model type that would be needed, provided it could be trained efficiently. There are indications that small feedforward networks applied to concatenated input elements (concatenated one-hot tokens) are similarly capable to small transformers for language-based sequence classification tasks, suggesting that a sufficiently large pure feedforward model would effectively learn to generate natural language \citep{badger2022small}. These models would be inefficient for longer sequences with bigger vocabularies, however, as they scale with $O(d^2tn)$ in parameter number with the length of the input $n$ and token size $t$ and hidden dimension $d$. 

    In this work we show that as a general rule, attention is useful for language tasks that would not be expected to require a globally invertible function, but this transformation does not appear to be useful for tasks requiring invertibility (autoencoding and retrieval in particular).

    \subsection{How many parameters are necessary for language modeling?}

    The TinyStories dataset was introduced with the goal of testing how large a transformer model and in particular what hidden layer size is required in order to accurately model a limited subset of the English language \citep{eldan2023tiny}. In this work the most efficient transformer alternatives have generally benefited from a larger hidden layers than transformers, which leads to the question: model architecture aside, how many parameters are required for language modeling? 

    To start to answer this question, we can consider the language task of sentence completion. Suppose there were a huge number of valid English sentences, perhaps $m = 10^{570}$ as an upper bound. Without knowing how to model these sentences, we can view them as unique points in an arbitrarily high-dimension space and apply a result from the concentration of measure phenomenon to determine the number of dimensions required to accurately. The Johnson-Lindenstrauss lemma provides us with the result that the same $m$ points may be represented with arbitrary precision in a space that is on the order of $8 \log m = 8 \ln 10^{570} \approx 1312 $ dimensional. More precisely, this lemma states that for some small $\epsilon > 0$ for set $X$ of $m$ points in $\Bbb R^N$, for when

    \begin{equation}
    n > 8\ln (m) / \epsilon^2
    \label{eq6}
    \end{equation}
    
    there is a linear representation map $f: \Bbb R^N \to \Bbb R^n$ such that for all $u, v \in X$ the following inequality holds:
    
    \begin{equation}
    (1 - \epsilon) ||u - v||_2 \leq ||f(u) - f(v)||_2 \leq (1 + \epsilon)||u - v||_2
    \label{eq7}
    \end{equation}

    In words, (\ref{eq7}) states that there exist a linear map $f$ with dimensionality $n$ that approximates $X$ of dimension $N$ arbitrarily well according to the $L^2$ distance between any two points in both sets, and the lower bound of the smaller dimension $n$ is governed by (\ref{eq6}).
    
    For more concrete estimates, consider the causal language modeling datasets used to train models to predict every `next' token in a concatenated sequence of tokens. This means that for a training dataset of $n$ tokens, the task is approximate $n$ points in some presumably very high-dimensional space. The FineWeb-edu 10BT dataset would thus require a model of dimension $8 \ln (10 \times 10^{9}) \approx 8 * 23 = 184$ for perfect memorization, and this number grows extremely slowly as the dataset size increases. One of the largest datasets that open-source models are currently trained on is 15 trillion tokens (for Llama-3 \citep{dubey2024llama}), and in this case we can represent this dataset arbitrarily well in a space approximately $8 \ln (15 \times 10^{12}) \approx 8 * 30 = 240$ dimensional. The goal of language models is to generalize to larger datasets and thus we would hope the model would accurately predict the next tokens of a much larger dataset. But even supposing that this 15 trillion token dataset is only one millionth of the size of this generalized dataset, one would still only require a space of dimension $8 \ln (15 \times 10^{18}) \approx 8 * 44 = 352$.
    
    How does this relate to the number of parameters necessary in a model? Parameter number may be equivalent to the dimensionality of the model with respect to the points it can approximate, or else the model's `width' or hidden layer dimension is equivalent to the model's dimensionality. Because the vector space of successive layers of practically every modern deep learning model are dependent (activations in layer $n+1$ depend on layer $n$ and perhaps layer $n-1$ etc.) and dimensionality is defined on linear independence, it seems more likely that a model's dimensionality best corresponds with its hidden layer dimension. If this is true then a model of no more than around 1300 hidden layer elements should be capable of completing any English language sentence, or a model with a width of 350 can accurately predict a next token for a massive dataset. If these hidden widths were used for default Llama architectures, the resulting models would are around 100 million (given a hidden width of $d_m=350$) and 995 million (for $d_m=1300$) parameters.

\bibliographystyle{unsrtnat}
\bibliography{references}  

\beginsupplement
\section{Appendix}

All code for this paper is available on Github (\url{https://github.com/blbadger/maskedmixers}). The motivations and background for this work are explained in more detail in a series of posts, see for example \citep{badgermixer}. 

    \begin{figure}[h]
        \centering
        \includegraphics[width=0.7\textwidth]{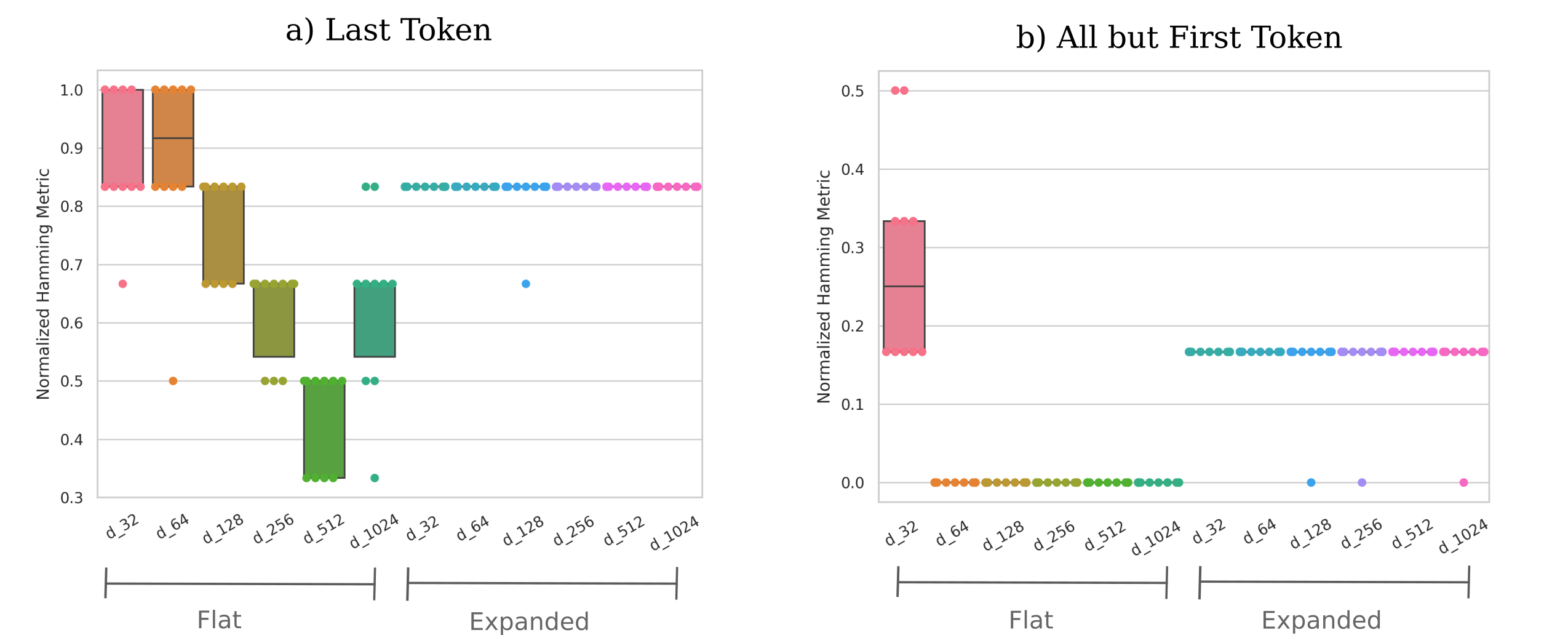}
        \caption{Flat masked mixers are more effective at passing information between tokens than expanded (e=2) masked mixers.}
        \label{figs3}
    \end{figure}
    
    \begin{figure}[h]
        \centering
        \includegraphics[width=0.7\textwidth]{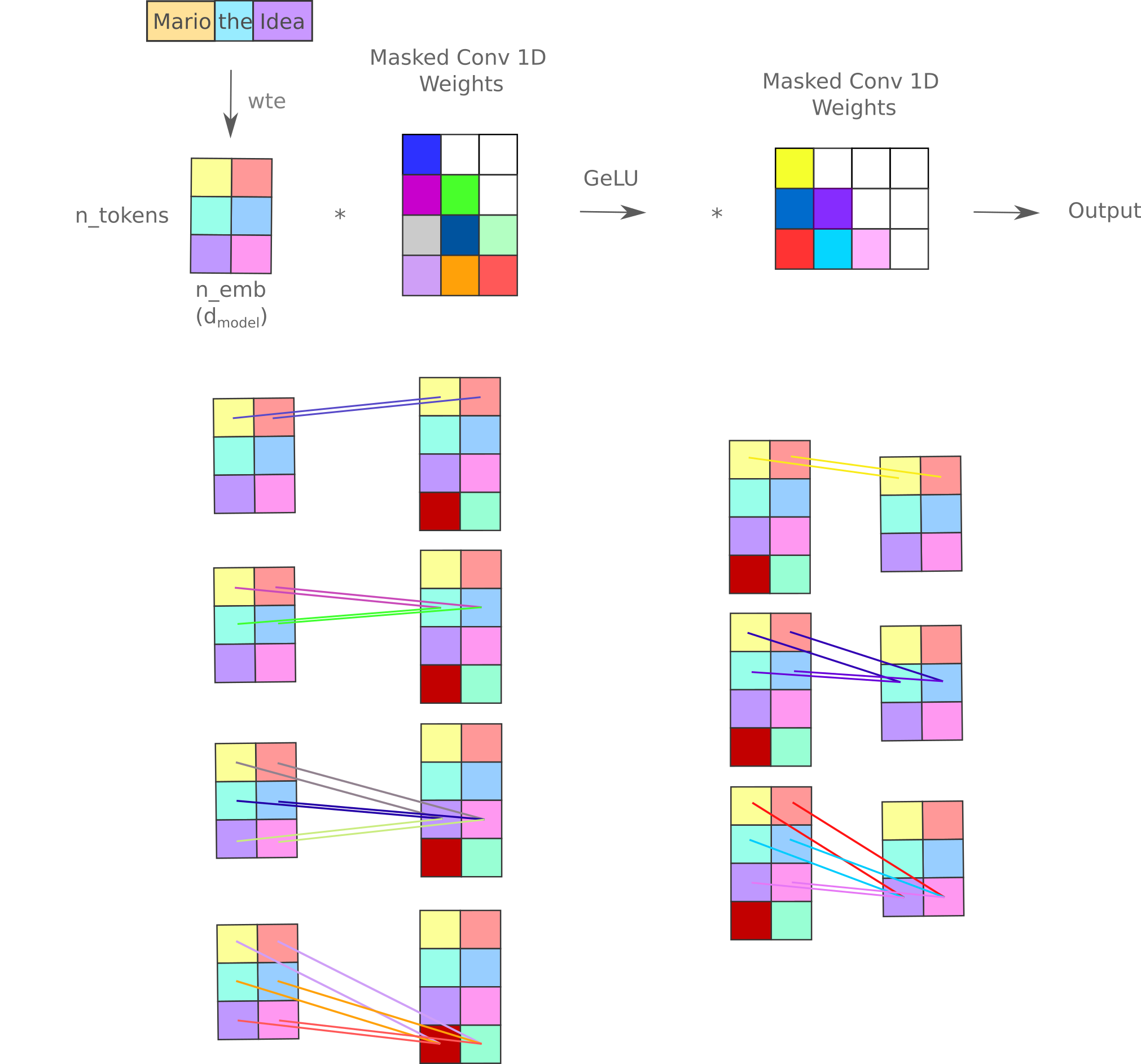}
        \caption{Mixers with expansion factors greater than one lead to unused parameters.}
        \label{figs1}
    \end{figure}

    \begin{figure}[h]
        \centering
        \includegraphics[width=0.7\textwidth]{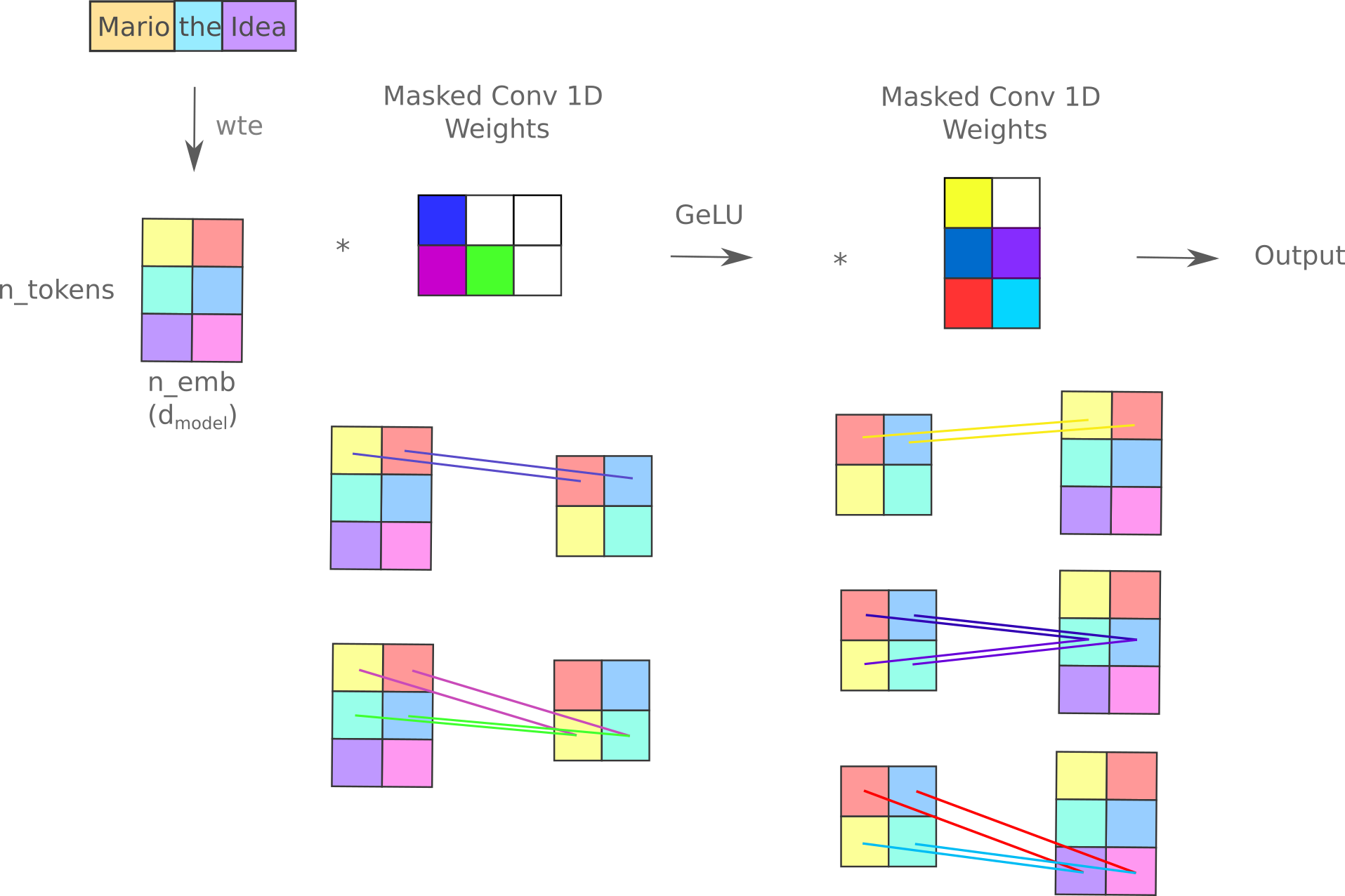}
        \caption{Mixers with expansion factors less than one lead to loss of information.}
        \label{figs2}
    \end{figure}

    \begin{figure}[h]
        \centering
        \includegraphics[width=0.99\textwidth]{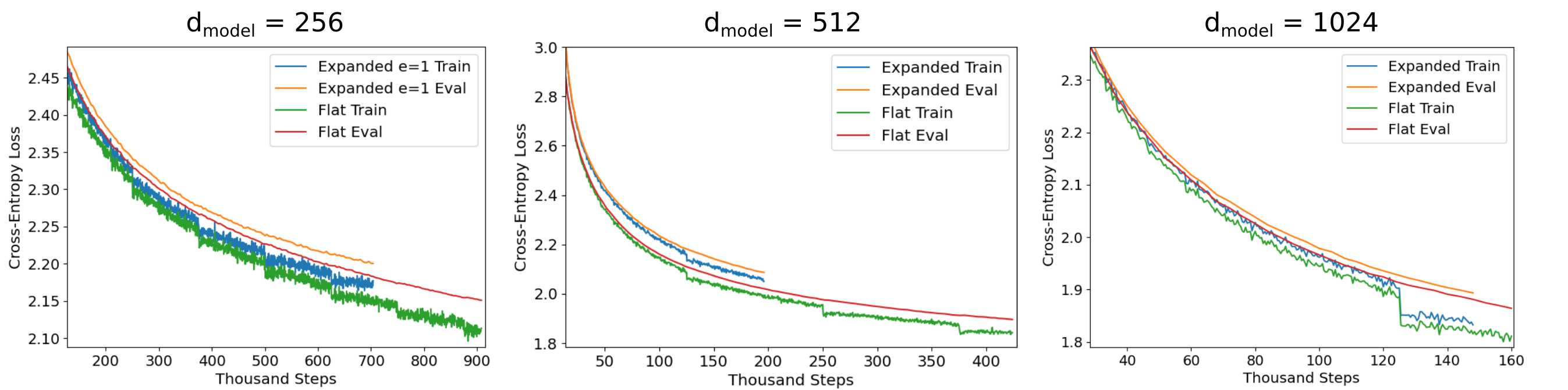}
        \caption{Flat mixers train more more efficiently than expanded ones.}
        \label{figs9}
    \end{figure}

    \begin{figure}[h]
        \centering
        \includegraphics[width=0.55\textwidth]{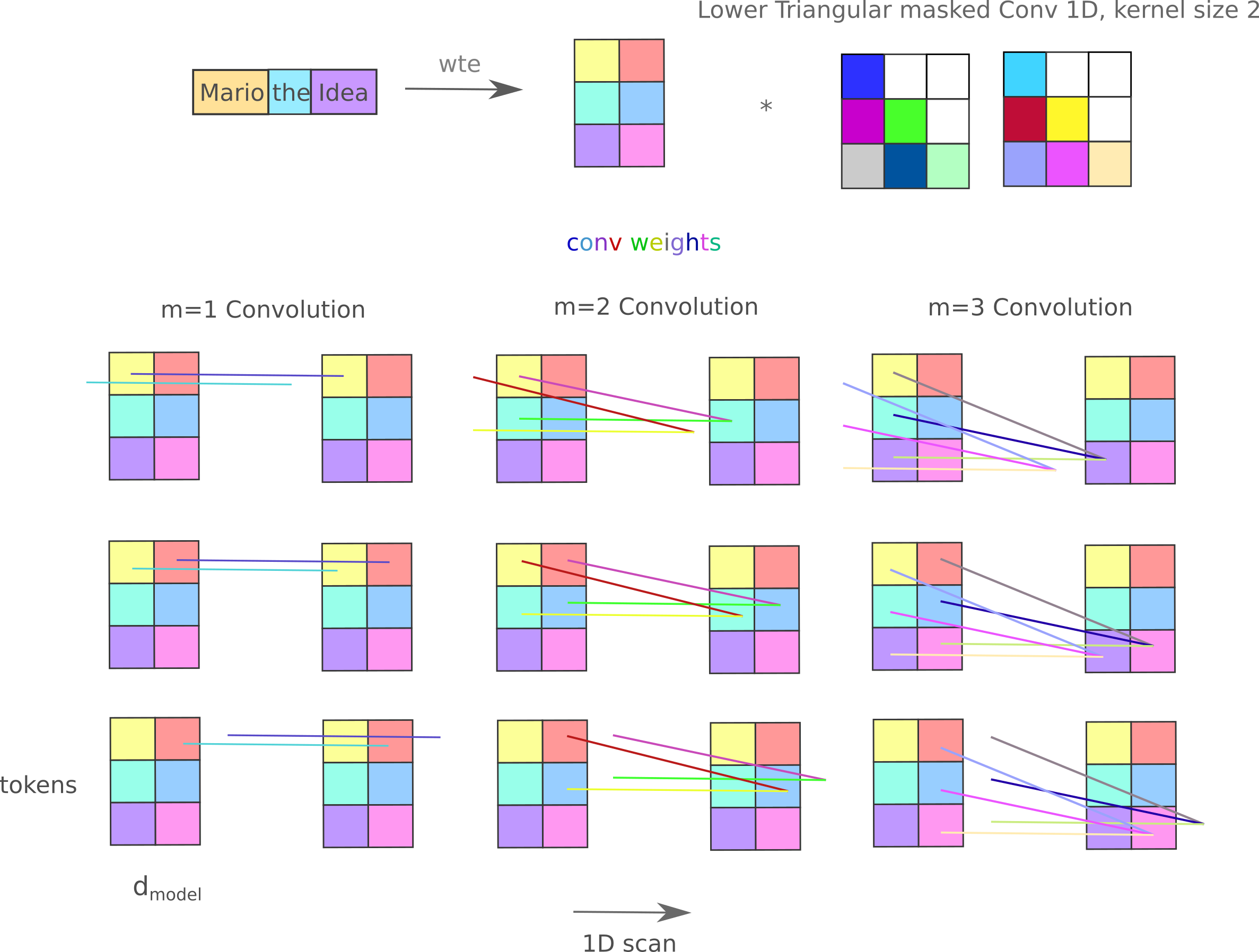}
        \caption{Convolutional kernel size 2 masked mixer with `same' style padding}
        \label{fig7}
    \end{figure}

        \begin{figure}[h]
        \centering
        \includegraphics[width=0.7\textwidth]{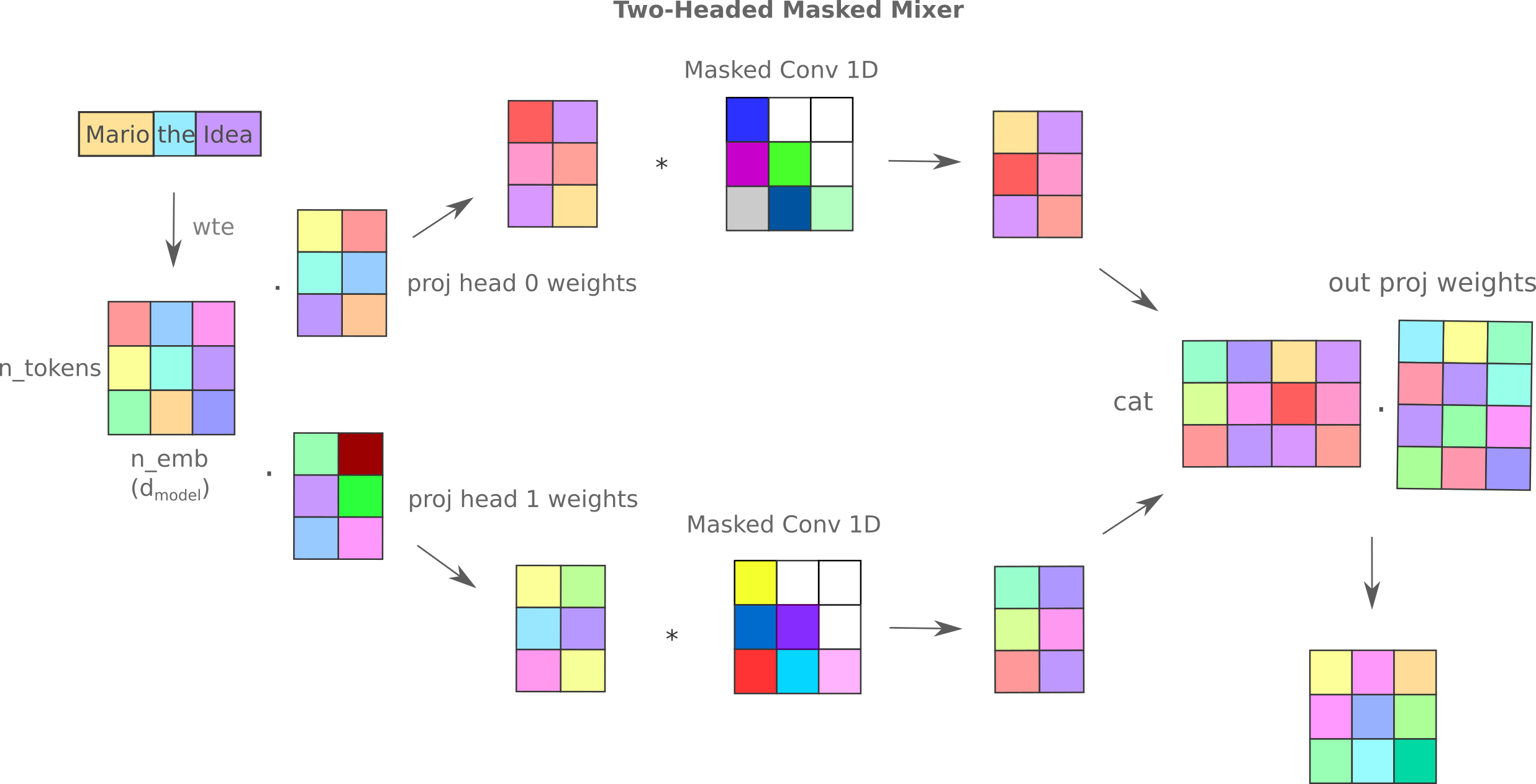}
        \caption{Multiheaded masked mixer architecture}
        \label{fig8}
    \end{figure}

    \begin{figure}[h]
        \centering
        \includegraphics[width=0.5\textwidth]{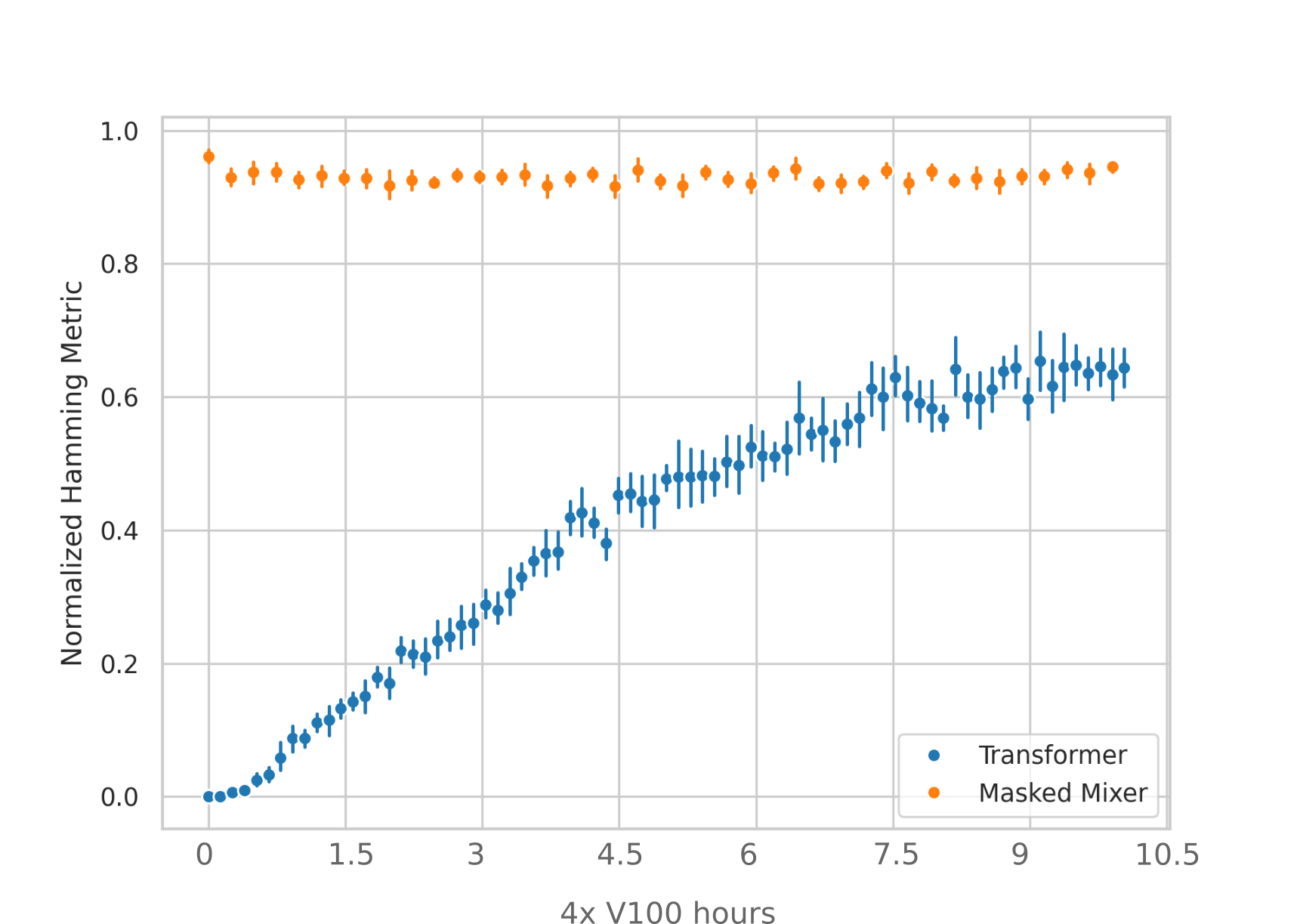}
        \caption{Normalized Hamming metric during extended TinyStories training (10h on 4x V100s) for 256-dim Llama and 512-dim masked mixer models.}
        \label{figs4}
    \end{figure}

    \begin{figure}[h]
        \centering
        \includegraphics[width=0.9\textwidth]{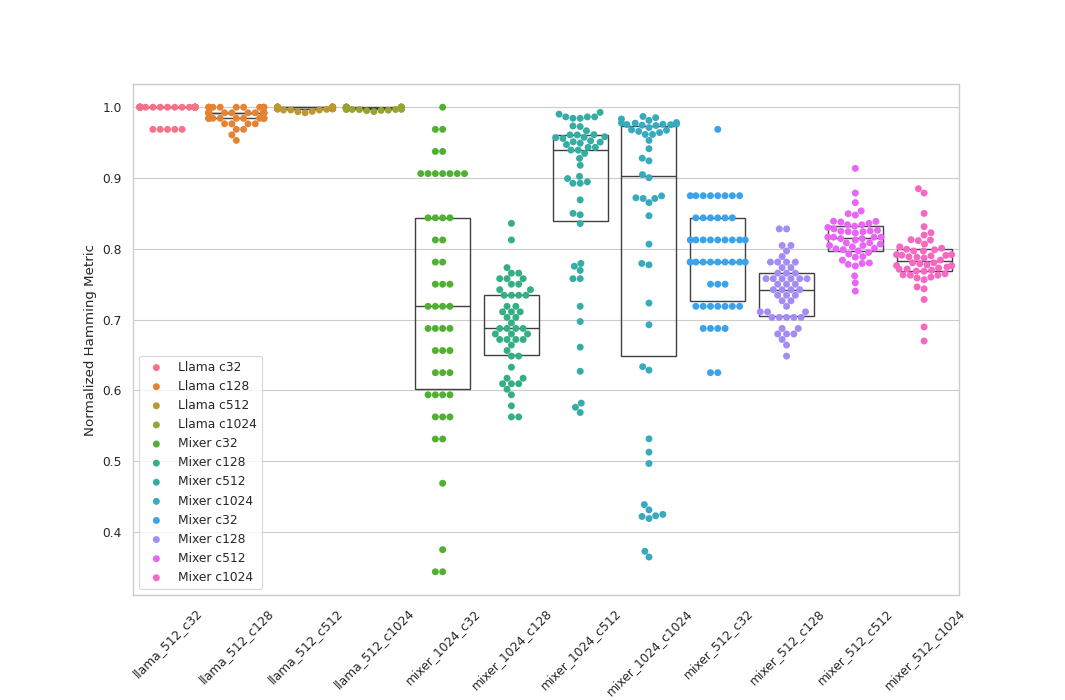}
        \caption{Normalized Hamming metric after training (20h on 4x V100s) for various models on FineWeb-edu 10BT, on random samples of that dataset.}
        \label{figs5}
    \end{figure}

    \begin{figure}[h]
        \centering
        \includegraphics[width=0.9\textwidth]{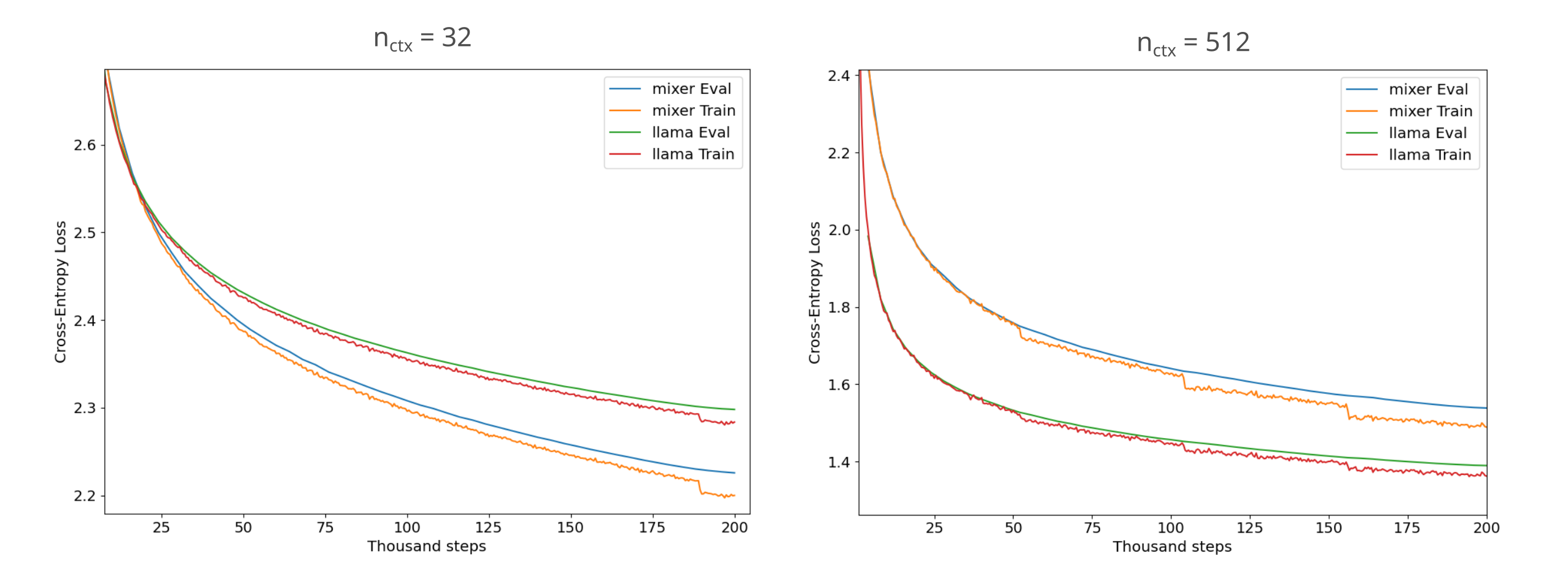}
        \caption{FineMath Causal Language Model Training.}
        \label{figs6}
    \end{figure}

    \begin{figure}[h]
        \centering
        \includegraphics[width=0.5\textwidth]{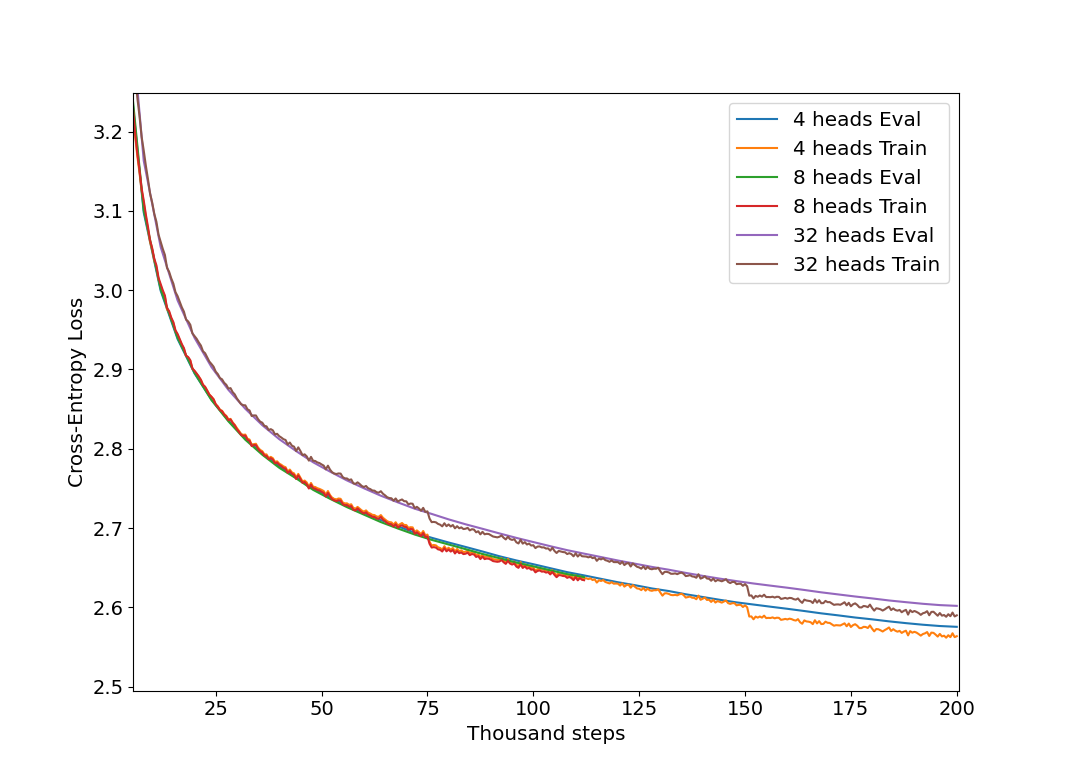}
        \caption{Llama model head number and FineWeb loss}
        \label{figs7}
    \end{figure}
    
    \begin{figure}[h]
        \centering
        \includegraphics[width=0.65\textwidth]{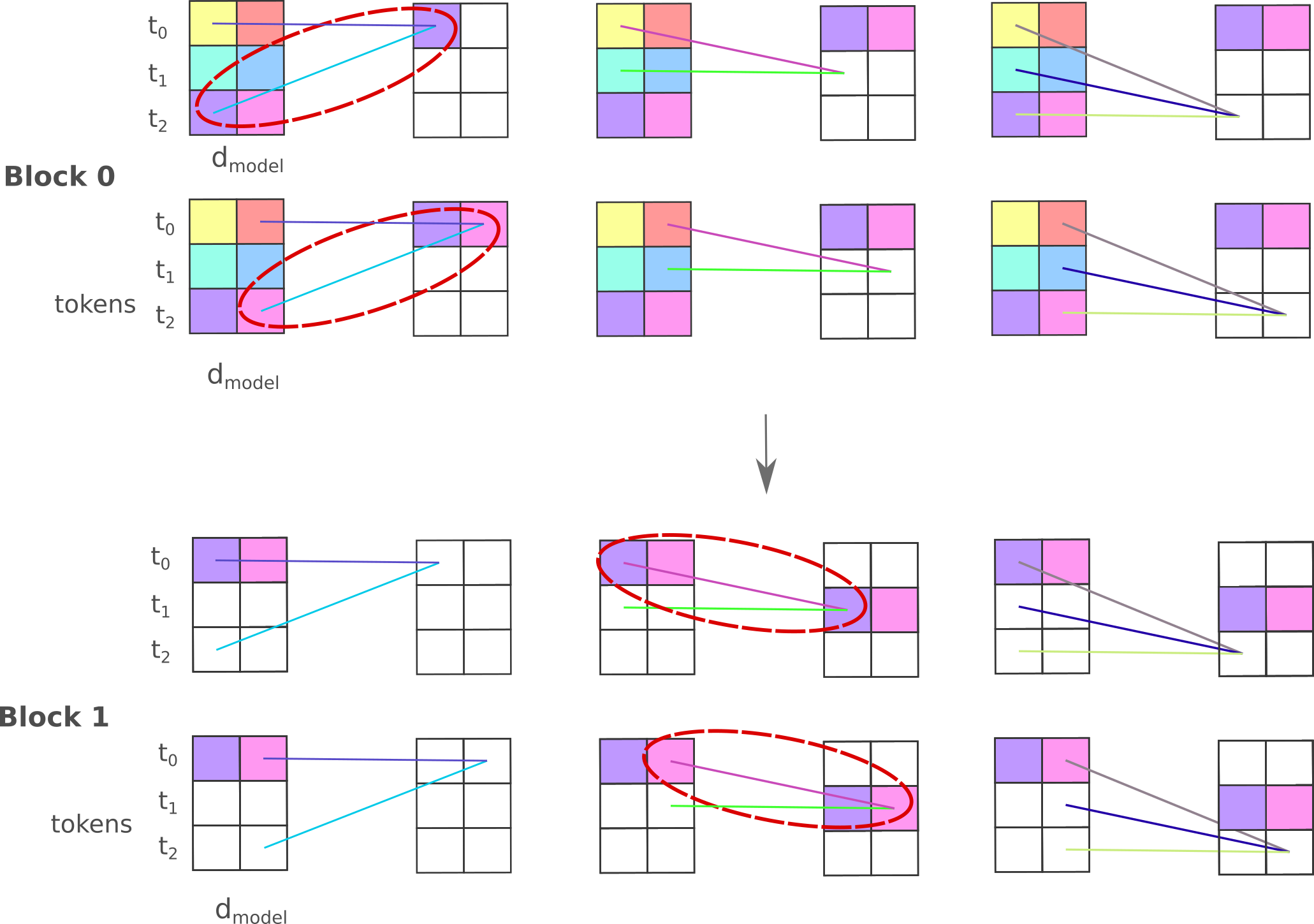}
        \caption{Bidirectional mixers must be modified to prevent `next' token information backfill (red dashed lines indicate flow).}
        \label{figs8}
    \end{figure}

\subsection{Representation accuracy-guided mixer architecture optimization}
    \label{representation_optim}

    A rough correlation is observed between representational accuracy and training efficiency for masked mixers: the original expanded-style architecture becomes more efficient as the hidden dimension increases, but the `flat' masked mixers with one convolutional layer between tokens which have better self- and non-self token representation are more efficient learners still (Figure \ref{figs9}, Table \ref{tables2}, \ref{tables3}). Both the increased efficiency with increased $d_{model}$ size as well as the increased efficiency of flat masked mixers relative to expanded ones correlate with increases in inter-token representation accuracy for these architectures (Figure \ref{figs3}). For Llama models the relationship between untrained or trained input accuracy and training efficiency is less clear (Table \ref{tables1}).

    It should be noted that the expansion factors for expanded masked mixers are in some sense superfluous hyperparameters when CLM masking is applied. This is because any expansion factors not equal to one leads to rectangular inter-token convolutional weight matrices after reshaping, and triangular masking these matrices leads to unused rows or columns (see Figure \ref{figs1} a graphical explanation of this phenomenon). This effectively makes expanded masked mixers with an expansion factor greater than one capable of viewing fewer samples per given time. But even when this effect is negated by considering losses at identical step numbers (rather than compute time) the expanded masked mixers are still slightly less efficient learners than `flat' masked mixers. For expansion factors less than one, triangular causal language masking results in the loss of input token information (see Figure \ref{figs2} for an example).  Because of the less accurate non-self token representation observed in the last section as well as the slightly worse training efficiency, we do not investigate the `expanded' mixer architecture further in this work.

    \subsection{Masked mixer training efficiency with multiple heads or non-unitary convolutional kernels}
    \label{multiheaded_mixers}

    Next we investigated whether increasing the number of inter-token parameters in the masked mixer would lead to more efficient training. As we had already found that flat mixers are more efficient TinyStories learners than expanded mixers, we started by testing the efficiency of masked mixers with parallel convolutions. Increasing the convolutional kernel's size leads to an increase in the number of inter-token parameters by a factor of the kernel size without increasing the depth of the inter-token transformations. A depiction of how a convolutional kernel of size 2 acts in the masked mixer is portrayed in Figure \ref{fig7} for convenience. We found that masked mixers with a convolution size of 4 outperforms those with size 1 using the 3060 node (Table \ref{tables3}), but that there was little difference for the 4x V100 (Table \ref{tables5}).

    We also investigated whether or not the use of multiple masked convolutional heads would result in greater training efficiency. To do this, we project each input dimension into a certain number of heads and apply a 1D masked convolution on each head, before concatenating the outputs and projecting back to match the dimension of $d_{model}$ before the feedforward layers are applied as shown in Figure \ref{fig8}. The training efficiency of a two-headed mixer is very slightly superior to that of a single-headed mixer, and is on par with a mixer with convolutional kernel of size 4 (Table \ref{tables5}). Both for increases in kernel size as well as adding convolutional heads the per-step loss was lower than for the vanilla flat masked mixer, but as fewer steps were taken with fixed compute the overall losses remains nearly identical.

    It may be wondered whether or not an increase in focus in the masked mixer could lead to increased learning efficiency.  One may do this by Softmax-transforming (without temperature) the 1D convolutional weights before re-masking. Adding the Softmax turns out to leads to much worse training efficiency than if these weights are not transformed regardless of the number of steps taken, however, indicating that increasing the `focus' of the masked convolution via Softmax alone does not benefit training (Table \ref{tables5}). 

    \subsection{Masked Mixers learn causal language modeling more efficiently than non-optimized modern transformers and optimized early transformer implementations}
    \label{unoptimized_transformers}
    
\begin{figure}[h]
        \centering
        \includegraphics[width=0.95\textwidth]{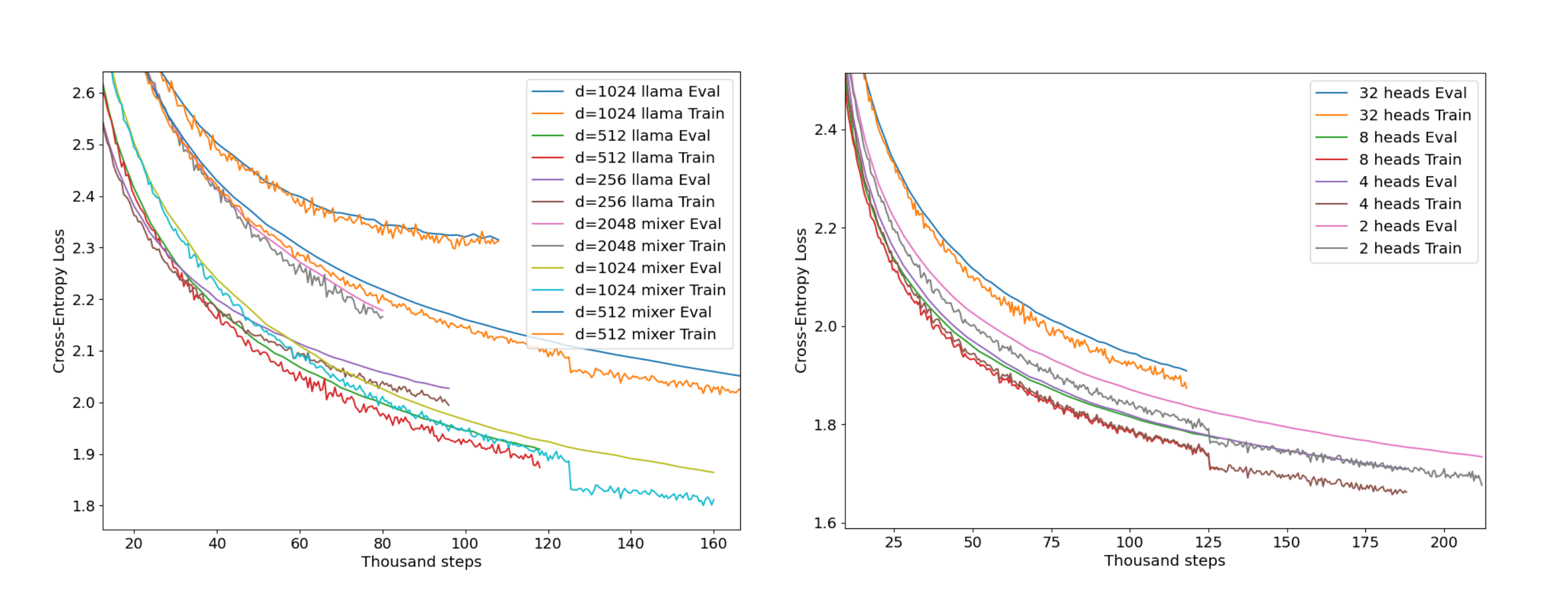}
        \caption{Left: Flat masked mixers are more efficient learners than modern transformers with a default attention head number, all with $b=16$ batch size except for $d_m=2048$ mixer and $d_m=1024$ llama which have $b=8$ to fit in memory. Right: Reducing the number of attention heads increases TinyStories training efficiency substantially.}
        \label{fig5}
    \end{figure}

\begin{figure}[h]
        \centering
        \includegraphics[width=0.95\textwidth]{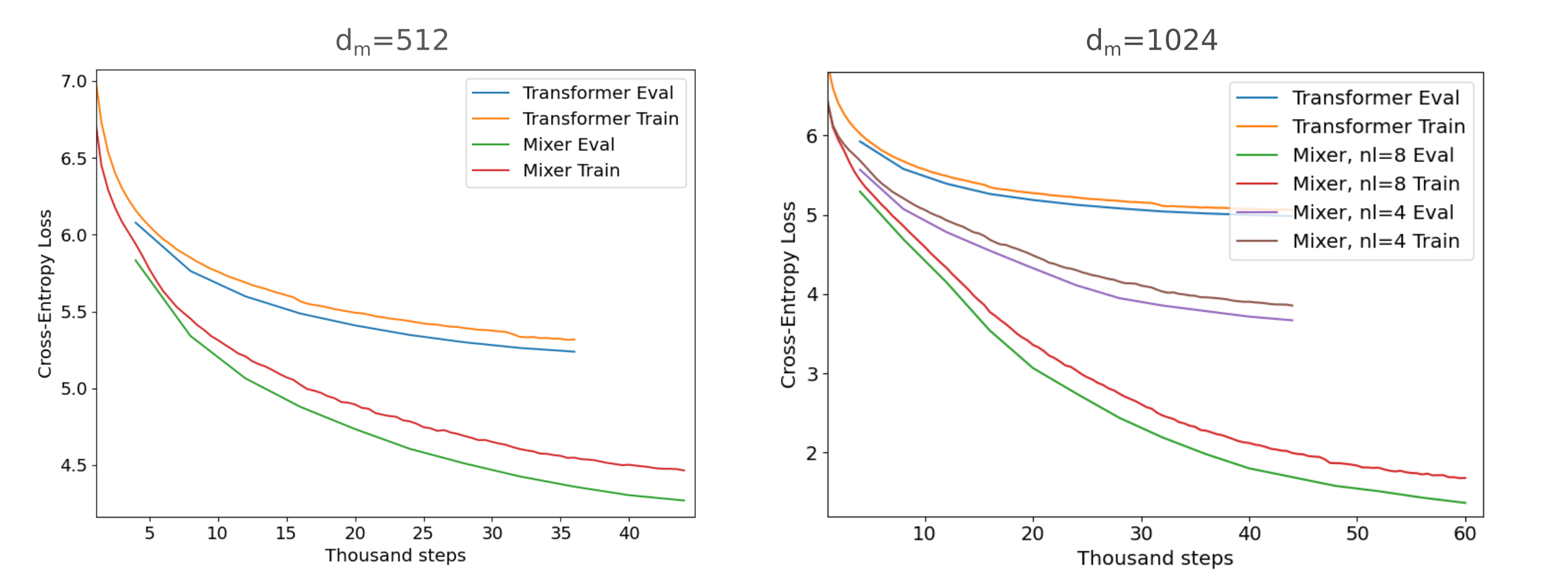}
        \caption{Masked mixer versus transformer (Llama-style encoder and decoder modules) autoencoder training loss on TinyStories for the specified $d_m$. On the right, note that the $n_l=8$ masked mixer is similar in compute and memory requirements to the transformer.}
        \label{figs10}
    \end{figure}

\begin{figure}[h]
        \centering
        \includegraphics[width=0.9\textwidth]{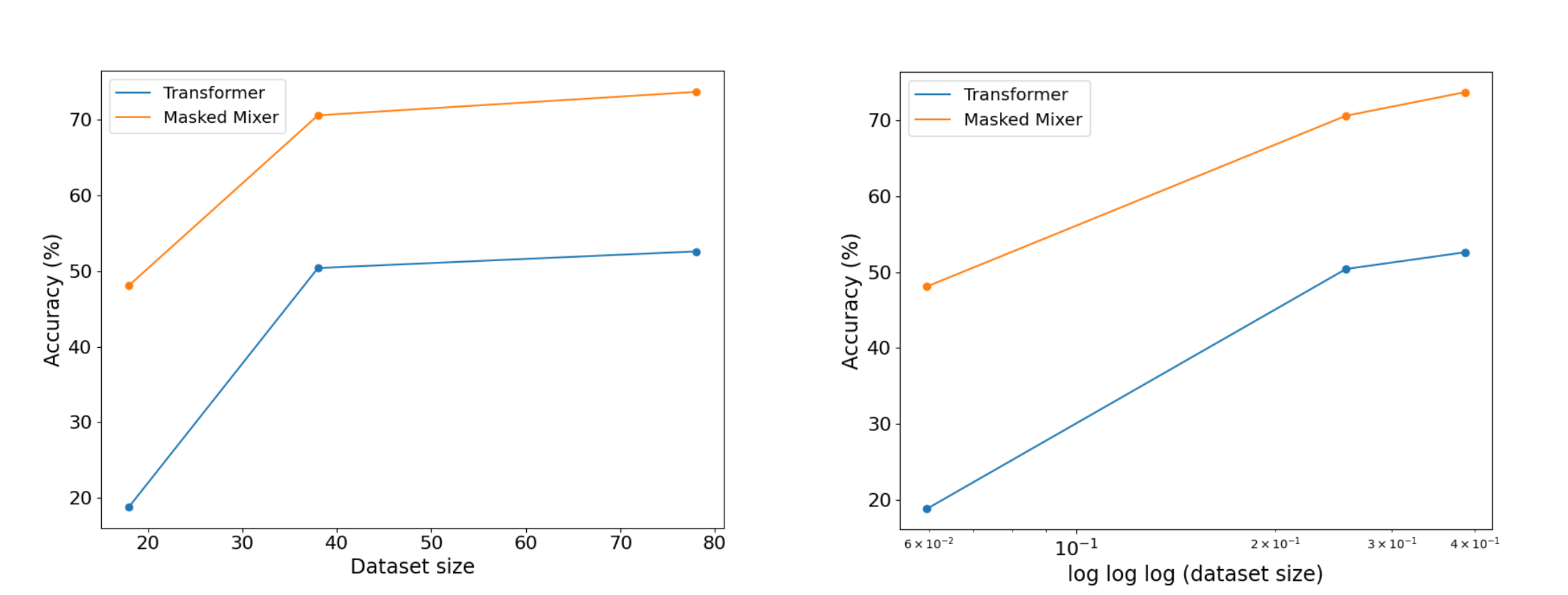}
        \caption{Masked mixer and transformer s=8192 top-1 accuracy with dataset size scaling.}
        \label{figs11}
    \end{figure}

    Flat masked mixers turn out to not only be more efficient TinyStories CLM learners than expanded mixers,  they are also more efficient than modern transformers (Llama-2 architecture) with default architectural hyperparameters, varying the $d_{model}$ after previously optimizing other hyperparameters such as the number of layers per model and batch sizes (Figure (\ref{fig5}, Table \ref{tables1}, Table \ref{tables3}). A comparison between flat masked mixer and transformer memory requirements for various context and $d_m$ sizes may be found in Table \ref{tables10} and Table \ref{tables11}.

    \subsection{Optimized versions of modern transformers are somewhat more efficient learners than optimized masked mixers}
    \label{optimized_transformers}

    Whilst optimizing the masked mixers and transformers in order to assess which model experiences a more efficient learning process, we focused on a number of hyperparameters: the hidden dimension $d_{model}$, the number of transformer or mixer block layers $n_l$, the AdamW learning rate, the number of attention heads and the batch size. For the Llama model we found that the most important single hyperparameter was the number of attention heads: decreasing this from 32 heads (the Llama 2 default) to 4 heads leads to a dramatic reduction in loss achieved per compute allocated, and somewhat surprisingly even on a per-step basis (Figure \ref{fig5}, Table \ref{tables4}). Decreasing the number of attention heads below 4 leads to worse performance, however. This fully optimized transformer learns to model TinyStories more efficiently than $d_m$-optimized masked mixers using the 3060 compute node (Tables \ref{tables3}, \ref{tables4}) but less efficiently on the 4x V100 node (Tables \ref{tables5}, \ref{tables6}) due to lower relative throughput for the transformer as these GPUs lack Flash Attention 2 \citep{dao2023flashattention2fasterattentionbetter}.
    
    \subsection{Compute and Training}
    
    In many cases it is a trivial task to introduce a new architecture with better scores on a benchmark if one has access to variable amounts of compute: simply increasing the compute used for training compared to the current approach, and ensure that the architecture has sufficient capacity to make use of an increase in compute. Because of this, we enforce constant compute in terms of clock time on fixed hardware (and software) on our model comparisons. 
    
    Early work in this paper (TinyStories training and input representation work) used a very modest amount of compute by modern standards, which for training was 12 hours of Nvidia RTX 3060 compute (with an i7-12700F CPU). After this was used to establish proofs of concept, we applied 2.25 hours with 4x Nvidia V100s (a Gigabyte T180-G20 with 2x 2680 E5 V4 CPUs) to the same dataset for CLM pretraining, and then scaled up the training to around 22 hours for datasets that are >10x the size of Tinystories, specifically FineWeb-edu 10BT and FineMath 4+ (around 10 billion tokens). Datasets were typically chunked (for $n_{ctx}<512$), padded, and tokenized before training for speed.
    
    Training on the 4x V100 cluster proceeded via Distributed Data Parallel. For TinyStories the 4x V100 cluster was power-limited to 200W per GPU with 877 memory and 1005 application clock speeds, which empirically decreased the cluster performance by a factor of around 15 percent. For modeling on the FineWeb and FineMath datasets the cluster was typically not power-limited and exact wall clock times were somewhat relaxed, swapped for architectures that were able to complete an exact step number within an approximate wall clock time. Retrieval training was performed on the 4x V100 cluster also using DDP. Training proceeded using AdamW optimizer with learning rates tuned to each model type ($\eta=5* 10^{-4}$ for masked mixers, $\eta=2*10^{-4}$ for transformers, $\eta=1*10^{-4}$ for all retrieval models) with linear decay and default betas. Loss masking was performed on pad tokens for both models, and attention masking was performed on pad tokens for transformers but not masked mixers.
    
    \subsection{Software infrastructure}
    
    Unless otherwise noted, for TinyStories the training runs in this study use either Hugging Face Transformers versions 4.36.0 (3060) or 4.41.1 (4x V100). All direct comparisons in this work are performed using identical versions of all major libraries (Pytorch, Transformers, Accelerate etc).  For training runs where Flash Attention 2 is noted as included, we use Transformers version 4.42.3 (FA2 was implemented in 4.42.0) but note that Flash Attention 2 is not available when using V100s (as it is only compatible with Ampere and newer architectures). For both 3060 and 4x V100 nodes we use Pytorch 2.3.1.

    \subsection{3060 compute fixed losses}
    
    Tables \ref{tables1}, \ref{tables2}, \ref{tables3}, and \ref{tables4} detail minimum losses achieved after 12 hours of compute with a single Nvidia RTX 3060 (12GB) on the first 2M samples of TInyStories. All models were trained using the `transformers.trainer` utility \citep{wolf-etal-2020-transformers} with evaluations every 4k steps and training statistics recorded every 500 steps. 
    
    \begin{center}
    \begin{table}
    \begin{center}
    \begin{tabular}{||c c c c c||} 
     \hline
       & $d_{model}=128$ & $d_{model}=256$ & $d_{model}=512$ & $d_{model}=1024$, b=8 \\ [0.5ex] 
     \hline\hline
      Train & 2.38 & 1.99 & 1.87 & 2.31  \\ 
     \hline
      Eval & 2.40 & 2.02 & 1.91 & 2.32  \\
     \hline
    \end{tabular}
    \end{center}
    \caption{Llama model training loss (n=8, h=32) for TinyStories on 12h 3060}
    \label{tables1}
    \end{table}
    \end{center}

    \begin{center}
    \begin{table}
    \begin{center}
    \begin{tabular}{||c c c c||} 
     \hline
       & $d_{model}=256, e=1$ & $d_{model}=512$ & $d_{model}=1024$ \\ [0.5ex] 
     \hline\hline
      Train & 2.17 & 2.05 & 1.83  \\ 
     \hline
      Eval & 2.20 & 2.08 & 1.89   \\
     \hline
    \end{tabular}
    \end{center}
    \caption{Expanded Mixer losses (e=2 unless otherwise noted) for TinyStories on 12h 3060}
    \label{tables2}
    \end{table}
    \end{center}

    \begin{center}
    \begin{table}
    \begin{center}
    \begin{tabular}{||c c c c c c||} 
     \hline
       & $d_{model}=256$ & $d_{model}=512$ & $d_{model}=1024$ & $d_{model}=2048$ & $d_{model}=1024, k=4$  \\ [0.5ex] 
     \hline\hline
      Train & 2.11 & 1.84 & 1.81 & 2.05 & 1.76 \\ 
     \hline
      Eval & 2.15 & 1.89 & 1.86 & 2.07 & 1.82 \\
     \hline
    \end{tabular}
    \end{center}
    \caption{Flat masked mixer losses for TinyStories on 12h 3060.}
    \label{tables3}
    \end{table}
    \end{center}

    \begin{center}
    \begin{table}
    \begin{center}
    \begin{tabular}{||c c c c c||} 
     \hline
       & $n_{heads}=32$ & $n_{heads}=8$ & $n_{heads}=4$ & $n_{heads}=2$ \\ [0.5ex] 
     \hline\hline
      Train  & 1.87 & 1.70 & 1.66 & 1.68  \\ 
     \hline
      Eval & 1.91 & 1.77 & 1.71 & 1.73  \\
     \hline
    \end{tabular}
    \end{center}
    \caption{Transformer Heads and losses for TinyStories for 12h 3060.}
    \label{tables4}
    \end{table}
    \end{center}

    \subsection{4x V100 fixed compute losses}
    
    Tables \ref{tables5} and \ref{tables6} give the fixed losses achieved after 2.25 hours on a 4x V100 (16GB) Gigabyte T180-G20 server node. The `transformers.trainer` utility of the transformers library \citep{wolf-etal-2020-transformers} was again used with 4k steps between evaluations and 500 steps between training statistic recordings.  Distributed Data parallel training was performed using the Hugging Face accelerate (\citep{accelerate}) integrations in the trainer utility, which wraps the Pytorch-native DDP utilities (\citep{NEURIPS2019_bdbca288}).
    
    \begin{center}
    \begin{table}
    \begin{center}
    \begin{tabular}{||c c c c c c||} 
     \hline
       & 1 head & 2 heads & 2 heads (softmax) & 1 head (k=4) & 1 head (k=4) and 2 wtes \\ [0.5ex] 
     \hline\hline
      Train  & 1.63 & 1.60 & 2.13 & 1.62 & 1.61  \\ 
     \hline
      Eval & 1.72 & 1.72 & 2.15 & 1.74 & 1.71 \\
     \hline
    \end{tabular}
    \end{center}
    \caption{Optimized Masked Mixer cross-entropy loss for TinyStories with 4x V100}
    \label{tables5}
    \end{table}
    \end{center}

    \begin{center}
    \begin{table}
    \begin{center}
    \begin{tabular}{||c c c c c||} 
     \hline
    & $b=16, \eta=0.02$ & $b=16$ & $b=32$ & GPT, $b=32$\\ [0.5ex] 
     \hline\hline
      Train & 1.78 & 1.76 & 1.68 & 1.82 \\ 
     \hline
      Eval  & 1.82 & 1.79 & 1.73 & 1.77 \\
     \hline
    \end{tabular}
    \end{center}
    \caption{4x V100 Optimized transformer model loss for TinyStories (no Flash attention 2, Llama style $d_{model}=512$, $n_l=8$, $\eta=0.005$, $n_h=4$ unless otherwise noted)}
    \label{tables6}
    \end{table}
    \end{center}

    \begin{center}
    \begin{table} 
    \begin{center}
    \begin{tabular}{||l c||} 
     \hline
      Pretraining Dataset & Top-1 Evaluation Accuracy (\%)  \\ [0.5ex] 
     \hline\hline
      FineWeb & 69.0 \\ 
      \hline
      FineWeb, autoencoder & 70.7 \\
     \hline
      FineWeb, 2 epochs InfoNCE & 71.3 \\
     \hline
     FineMath 4+ & \textbf{81.8} \\
     \hline
    \end{tabular}
    \end{center}
    \caption{FineMath retrieval evaluation top-1 accuracy for a $d_m=512, n_l=16$ masked mixer pretrained via next token prediction (unless otherwise noted) on the denoted dataset, followed by InfoNCE training for one epoch unless other wise noted.}
    \label{tables8}
    \end{table}
    \end{center}

    \begin{center}
    \begin{table}
    \begin{center}
    \renewcommand{\arraystretch}{1.2}
    \begin{tabular}{||l c c c c c c c c c||} 
     \hline
      Model & $n=32$ & 64 & 128 & 256 & 512 & 1024 & 2048 & 4096 & 8192 \\ [0.5ex]
     \hline\hline
      Transformer & 98.1 & 98.1 & 96.9 & 95.4 & 93.8 & 91.0 & 88.0 & 84.3 & 79.8 \\
      \hline
      Masked Mixer & 99.8 & 99.7 & 99.3 & 99.0 & 98.4 & 97.9 & 97.0 & 95.8 & 94.0 \\
     \hline
    \end{tabular}   
    \end{center}
    \vspace{0.1cm}
    \caption{FineWeb-edu retrieval top-1 accuracy (\%) for various sample sizes $n$, pretrained on FineMath 4+ and post-trained on s=180k synthetic retrieval samples from the same dataset. All models are $d_512, n_l=16$}
    \label{tables12}
    \end{table}
    \end{center}

\subsection{Memory scaling properties}

    The efficiency of the masked mixer relative to a transformer of the same number of trainable parameters is apparent as follows: if a masked mixer is compared to that for a transformer of the same width, we see that the masked mixer during training is between four and eight times as memory-efficient. Both models are $O(n^2)$ memory complexity, but the masked mixer has much lower constant factors as they typically have less inter-token parameters and more importantly (for training) many fewer effective parameters that require independent gradient values.

    \begin{center}
    \begin{table}
    \begin{center}
    \begin{tabular}{||c c c c c c||} 
     \hline
       & $n_c$=512 & $n_c$=1024 & $n_c$=2048 & $n_c$=4096 & $n_c$=8192 \\ [0.5ex] 
     \hline\hline
      $n_l$=4 & 2071 & 2341 & 2637 & 3573 & 6491  \\ 
     \hline
      $n_l$=8 & 2431 & 2869 & 3425 & 5111 & 10527  \\
     \hline
     $n_l$=16 & 2695 & 3159 & 3811 & 5879 & OOM  \\ 
     \hline
    \end{tabular}
    \end{center}
    \caption{Flat Masked Mixer memory requirements (MB) on an RTX 3060 (12GB), $b=16$, $n=8$ layers}
    \label{tables10}
    \end{table}
    \end{center}

    \begin{center}
    \begin{table}
    \begin{center}
    \begin{tabular}{||c c c c c c||} 
     \hline
         & $n_c$=512 & $n_c$=1024 & $n_c$=2048 & $n_c$=4096 & $n_c$=8192 \\[0.5ex] 
     \hline\hline
      $n_l$=4 & 2323 & 3275 & 6809 & OOM & OOM  \\ 
     \hline
      $n_l$=8 & 3176 & 4800 & 10126 & OOM & OOM  \\
     \hline
     $n_l$=16 & 4876 & 7750 & OOM & OOM & OOM  \\ 
     \hline
    \end{tabular}
    \end{center}
    \caption{Transformer memory requirements on an RTX 3060 (12GB) in MB ($n=8$ layers, $h=32$ attention heads, $b=16$ batch size unless otherwise noted)}
    \label{tables11}
    \end{table}
    \end{center}

    Note that the above tables do not capture the memory required for training with optimizers that must save multiple gradient values per trainable parameter, or for batched inputs multiple activation values. The smaller constant factors for memory complexity in masked mixers compared to transformers lead to even larger efficiency gains once these are accounted for.
    
\end{document}